\documentclass[11pt]{article}
\usepackage[margin=1in]{geometry}
\usepackage[T1]{fontenc}
\usepackage[utf8]{inputenc}
\usepackage{lmodern}
\usepackage{microtype}
\usepackage{setspace}
\usepackage{amsmath,amssymb}
\usepackage{graphicx}
\usepackage{booktabs}
\usepackage{xurl}
\usepackage[round,authoryear]{natbib}
\setcitestyle{aysep={,},citesep={;}}
\usepackage[hidelinks]{hyperref}
\newcommand{\doi}[1]{doi: \href{https://doi.org/#1}{\nolinkurl{#1}}}
\usepackage{enumitem}
\usepackage{caption}
\DeclareCaptionLabelFormat{panelb}{#1 #2(b)}
\usepackage{titling}
\pretitle{\begin{center}\fontsize{19}{23}\selectfont\bfseries}
\posttitle{\par\end{center}\vskip 0.75em}
\preauthor{\begin{center}\large}
\postauthor{\par\end{center}}
\renewenvironment{abstract}
  {\begin{center}\bfseries\large Abstract\end{center}\begin{quotation}\normalsize}
  {\end{quotation}}
\title{Comparing Object Detection Models for Electrical Substation Component Mapping}
\author{Namish Bansal\textsuperscript{3,*}, Haley Mody\textsuperscript{3,*},\\
Dennies Kiprono Bor\textsuperscript{1,2}, Dante Groccia\textsuperscript{1}, and Edward J. Oughton\textsuperscript{1}\\[0.5em]
\normalsize\textsuperscript{1}Department of Geography and Geoinformation Science,\\
\normalsize College of Science, George Mason University, Fairfax, VA, USA\\
\normalsize\textsuperscript{2}Space Weather Lab, George Mason University, Fairfax, VA, USA\\
\normalsize\textsuperscript{3}Aspiring Scientists Summer Internship Program (ASSIP),\\
\normalsize College of Science, George Mason University, Fairfax, VA, USA\\[0.5em]
\normalsize\textsuperscript{*}Namish Bansal and Haley Mody contributed equally to this work\\
\normalsize and share first authorship.}
\date{}

\begin{document}
\maketitle

\begin{abstract}
Electrical substations are a significant component of an electrical grid. Indeed, the assets at these substations (e.g., transformers) are vulnerable to hazards such as hurricanes, flooding, earthquakes, and geomagnetically induced currents (GICs). Because failures can have significant economic and public safety implications, identifying key substation components is essential for quantifying vulnerability. Unfortunately, traditional manual mapping of substation infrastructure is time-consuming and labor-intensive. Therefore, an autonomous solution utilizing computer vision models is preferable, as it offers greater convenience and efficiency. In this study, we train and compare 16 models on a manually labeled dataset of US substation images. These models include 12 You Only Look Once (YOLO) models, 2 Roboflow Detection Transformer (RF-DETR) models, and 2 Cascade R-CNN models. RF-DETR-large achieved the highest overall detection performance with mAP@50 and mAP@50:95 scores of 0.881 and 0.632, respectively. Across all models, alternate energy systems were detected most accurately, while transformers and reactors were more difficult to identify due to their smaller size and greater visual variability. Applying our best-performing model to nationwide imagery yielded approximately 22,591 component detections across 11,083 unique substations within the United States. These detections were broken down by state and Federal Energy Regulatory Commission (FERC) regions, with Florida (2,478 detections) and Midcontinent Independent System Operator (MISO; 4,329 detections) having the largest number of detections in their respective categories.
\end{abstract}
\clearpage

\section{Introduction}

Modern power grid operations depend on a reliable and efficient electrical grid. \citep{erenoglu2024power,xu2025quantifying,sundar2023meteorological,xu2024resilience,kalt2021global}. The global electrical grid spans millions of miles and supplies power to billions of people \citep{kalt2021global}. Electricity is generated at power plants, transmitted over long distances, and distributed at lower voltages \citep{imdadullah2021electric,ohanu2024comprehensive}. Key nodes in enabling long-distance energy transmission are electrical substations, which either step up or step down voltage levels \citep{yao2023site}. They are pivotal in ensuring that electricity is distributed to consumers and businesses.

However, the importance of substations and their components goes beyond standard maintenance and regulation. Because the electrical grid is considered critical national infrastructure, any failure can have significant economic and public safety impacts. In recent years, the grid has become increasingly vulnerable to hazards such as geomagnetically induced currents (GICs), hurricanes, flooding, and earthquakes, all of which can damage substation components \citep{cmap,xu2025quantifying,xu2024resilience,erenoglu2024power,kaya2025downscaling}. To address these challenges, it is crucial to precisely identify the components in each substation and their respective locations, enabling risk assessments. Efforts have been made to map substation components across the country manually. Manually mapping out substations is not a feasible solution due to the sheer volume of the grid and its labor-intensive and time-consuming nature. Additionally, the fragmented nature of the grid leads to inconsistent data collection and labeling practices across regions, making consistency and standardization difficult across datasets \citep{meyur2022ensembles}.

For these reasons, an automated approach to substation component mapping is a more efficient and consistent solution \citep{usery2022geoai}. Our study investigates whether state-of-the-art computer vision models can be trained to identify and distinguish key substation components in high-resolution imagery accurately \citep{nguyen2019intelligent,liu2020research,siddiqui2020drone,han2020electrical,zheng2021component}. This enables substations to be mapped consistently and rapidly across the grid, reducing reliance on manual labor while improving data accuracy and timeliness. Ultimately, this enhances our ability to assess risk and respond to hazards at a national scale.

In this paper, we will explore various computer vision models that can be used for this task and attempt to predict the number of different substation components, using the United States (U.S.) as a case study. This poses the following research questions:

\begin{enumerate}[
    label=\arabic*.,
    align=left,
    leftmargin=0pt,
    labelindent=0pt,
    itemindent=2em,
    labelwidth=1.5em,
    labelsep=0.5em
]
\item \textit{What is the best object detection model that can be utilized to identify and differentiate between substation components effectively?}

\item \textit{What components are most accurately detected and why?}

\item \textit{Can we effectively count the total number of nationwide substation components in the United States and analyze them by both state and Federal Energy Regulatory Commission (FERC) Order 1000 planning regions?}
\end{enumerate}

In Section 2, a literature review will be conducted to compare how various research papers approach object detection for power infrastructure modeling. Section 3 explores the methodology employed for model training, including data preparation, model selection, and implementation. Section 4 details the results of the model training, application of techniques, and the inference performed. Section 5 analyzes these results and ties them back to the aforementioned research questions. Finally, Section 6 concludes the study by summarizing key findings and discussing their broader implications.

\clearpage
\section{Literature Review}

\begin{figure}[htbp]
\centering
\includegraphics[width=\textwidth,height=0.78\textheight,keepaspectratio]{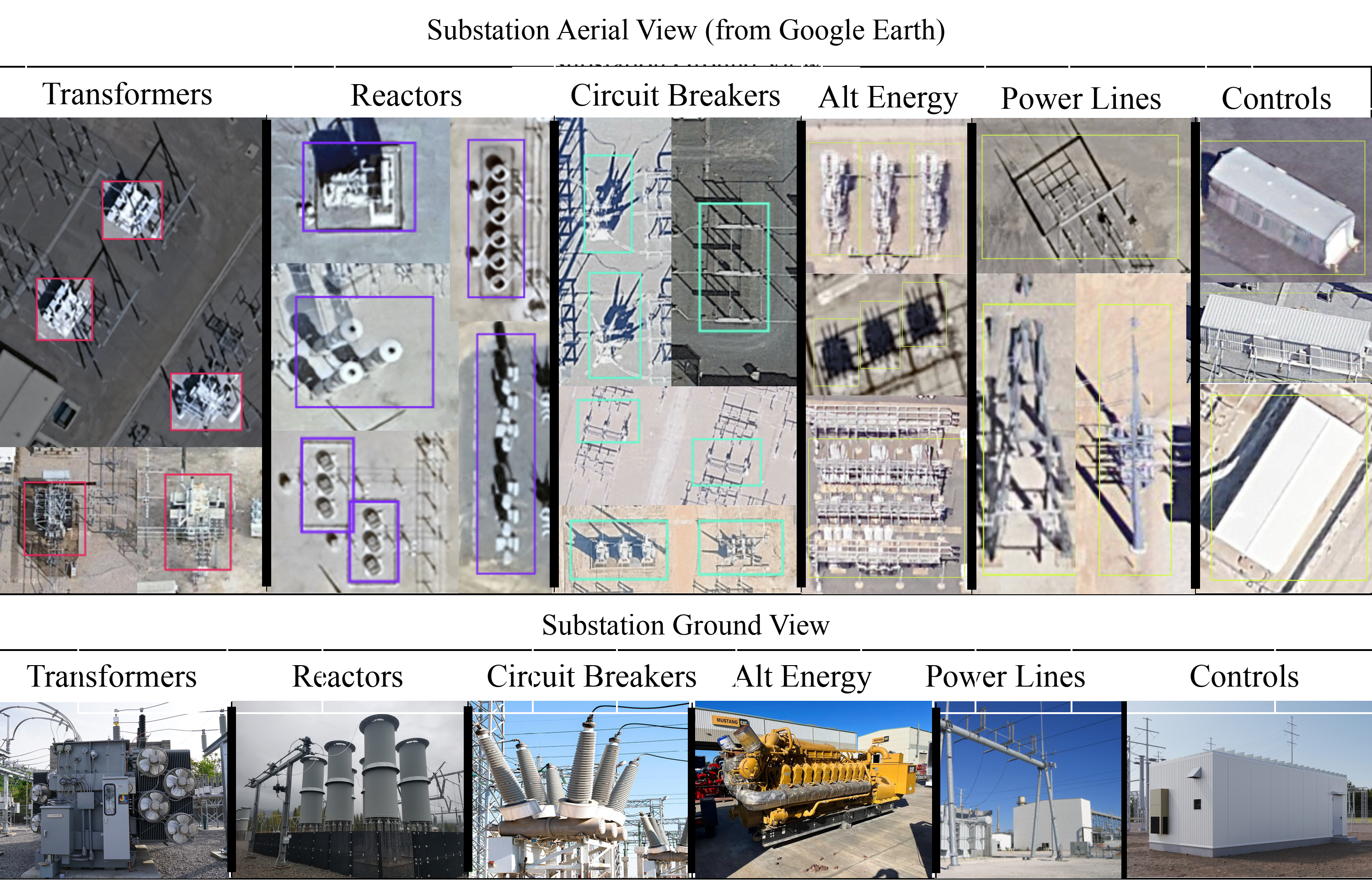}
\caption{Annotation samples of different substation components.}
\label{fig:components}
\end{figure}

As modern society's reliance on energy supply increases, protecting critical infrastructure, such as power infrastructure, from natural hazards is becoming increasingly important. Substations contain various components, as shown in Figure (\ref{fig:components}).

Substation infrastructure may be affected by failures such as overload or overvoltage \citep{zhang2023condition}. Substation monitoring is critical, as infrastructure failures can severely impact public safety and local wildlife. Traditional manual inspections are time-consuming and costly, underscoring the need for automated solutions based on artificial intelligence and computer vision \citep{lavado2025enhancing,liu2026accurate,lekidis2022electricity,kaur2024developments,tiede2021mask}.

The current research in computer vision is dominated by the You Only Look Once (YOLO) family \citep{hussain2023yolov1}. Recent versions of YOLOv10 and YOLOv11 typically achieve lower latency and higher mean average precision (mAP) across aerial and remote sensing benchmarks \citep{chi2024improved,yang2025yolov10msa,chandra2025novel}. Due to this trade-off between inference speed and detection, YOLO models are often used for power grid substation detection and insulator inspection workflows \citep{faisal2025deep,hu2025uav}. Compared with newer paradigms such as detection transformers (DETR), YOLO still achieves faster inference without sacrificing accuracy or precision \citep{mattern2025comparison,sapkota2025rfdetr}. Furthermore, custom YOLO modifications have been successfully adapted to high-resolution optical satellite imagery for mapping electric pylons across wide geographic areas \citep{chi2024improved,tasyurek2024dgag} as well as handling severe visual obstructions like fog in overhead transmission corridors \citep{zhang2022finet}.

Studies have also fused the YOLOv8 algorithm with an enhanced small-object detection head to improve accuracy in unexpected environments \citep{li2025enhanced,aitelhaj2025apfyolov8,khalili2024sodyolov8,nie2024lightweight,ni2024smallobject}. Another study focuses exclusively on insulators and proposes a dynamically adaptive YOLOv8 model \citep{hu2025uav}. This model, based on YOLOv8, is modified to improve functionality regarding elliptical features. This approach demonstrates significant advances in research methods, accounting for unique challenges, and enabling a more comprehensive evaluation of object detection performance. However, it highlights a critical limitation in current research on power infrastructure modeling: it focuses solely on YOLO models and does not compare them with other model types. In particular, there is insufficient information about how YOLO models' training speed compares with that of other models. Due to the lack of comparison, it is difficult to determine which model provides the best detection.

Furthermore, the reviewed literature highlights dataset inconsistencies that directly affect mAP values. Dataset sizes vary drastically across the observed studies, which skews comparisons of algorithms and makes it difficult to standardize a benchmark and determine which models actually perform better than others \citep{kuznetsova2020open,padilla2021comparative,sowrirajan2020mococxr,yu2019scale,zhao2023detrs}

For instance, one study utilized 7,605 Red, Green, and Blue (RGB) images captured by Unmanned Aerial Vehicles (UAVs) to train a Single Shot MultiBox Detector algorithm (SSD) to detect porcelain and composite insulators, achieving an accuracy of 91-94\% for porcelain insulators and 87\% for composite insulators, despite using the same dataset images and training algorithm \citep{miao2019insulator}. By contrast, another study conducted in the same year used 3,500 RGB images with the YOLOv2 algorithm to detect glass insulators, achieving an accuracy of 88\%, nearly the same as the previous study, albeit with only about half as many images \citep{lai2019automatic}. These examples illustrate that often a larger dataset may result in higher accuracy, but that is not always the case. Variations in data collection methods, imaging conditions, and algorithm choices can all influence results \citep{bao2022defect,chen2023insuyolo}. Therefore, it may be harder to conclude which algorithm is more optimal for specific conditions \citep{zhang2023lightweight}. It is important to maintain a consistent and sufficiently large dataset to ensure meaningful model evaluations and minimize confounding variables that may skew the results of the data \citep{abdelfattah2020ttpla,vieiraesilva2023insplad}.

A newly released model, Roboflow Detection Transformer (RF-DETR), is another focus of our research paper, due to the limited number of studies evaluating its performance \citep{robinson2025rfdetr}. Building on DETR, RF-DETR incorporates deformable attention mechanisms, alongside self-supervised pretrained backbones derived from self-distillation with no labels (DINO) to achieve high-precision detection \citep{carion2020endtoend,zhu2020deformable,oquab2023dinov2}. In one recent study, the RF-DETR beat the YOLOv12 model in precision, achieving the highest of 0.946 in comparison to the extra-large (X), large (L), and nano (N) YOLOv12 variants with precisions of 0.93 each in single-class detection. Beyond single-class detection, RF-DETR also achieved higher precision in multi-class detection, with a mean average precision at an intersection over union (IoU) threshold of 0.50 (mAP@50) of 0.830, compared to the other models. The achieved efficiency in distinguishing between classes that need to be detected and obstacles makes it optimal for component localization under high spatial density \citep{sapkota2025rfdetr}. Recent transformer architectures such as RF-DETR, Fine-grained Distribution Refinement (D-FINE), and Light-Weight Detection Transformer (LW-DETR) have established new Pareto bounds over convolutional neural network (CNN)-based YOLO models on general benchmarks like Common Objects in Context (COCO) and Roboflow 100 for vision-language models (RF100-VL) \citep{peng2024dfine,chen2024lwdetr,robinson2025rfdetr}.

Cascade-based detectors have also occasionally been applied to substation equipment monitoring. One study proposed an infrared image recognition algorithm for power transmission and substation equipment built on an improved Cascade region-based convolutional neural network (R-CNN) framework for automatic recognition of electrical devices such as transformers and switches in substation environments \citep{li2026frnet}. Additionally, on the Northeastern University surface-defect detection (NEU-DET) dataset, the standard Cascade R-CNN model was used as a comparison point to an enhanced version; the mAP increased to 75\%, an increase of almost 9.2 percentage points over the unmodified Cascade R-CNN. Similarly, in another evaluation of defect detectors for photovoltaic equipment components, Cascade R-CNN was one of the eight benchmarked models. While it successfully detected all defect instances present, it showed reduced bounding-box localization precision relative to newer detectors, at times splitting a single defect into two separate detections \citep{jia2024defect}. However, studies relating to Cascade-based models detecting electrical substation components are often rare; our study aims to address this gap by comparing Cascade R-CNN against YOLO and RF-DETR models.

In addition to dataset size, environmental conditions often pose challenges for reliable object detection. Some types of visual obstructions can include fog, smoke, mist, or even blurry images. In a past study \citep{zhang2022finet}, the Synthetic Foggy Insulator Dataset (SFID) was utilized to introduce synthetic fog and assess detection performance under environmental obstacles. They created a new model called Foggy Insulator Network (FINet), built on the YOLOv5 framework, enhanced with a channel attention mechanism to detect infrastructure on the SFID.

To ensure a fair evaluation, all models were trained and tested using a single, consistent dataset, minimizing discrepancies when comparing precision and efficiency of different models. However, to fully evaluate models, future research should account for these data consistency and bias issues.

To include a more comprehensive review of power substation research, existing studies only evaluate a single model or a single model family \citep{yigit2025deep}. For example, a 2024 study \citep{he2024comprehensive} conducted a comparison of five YOLO generations (YOLO versions 5, 8, 9, 10, 11) on a power equipment detection dataset. The results showed mAP scores of 54\%, 56\%, 44\%, 48\%, and 57\%, respectively, with YOLOv11 achieving the highest detection performance. However, the study was limited to the YOLO family only and did not include transformer-based or two-stage detectors, failing to compare whether YOLO could truly outperform other model families.

Furthermore, work that is focused on aerial power infrastructure detection has tended to evaluate models in isolation rather than comparatively. A previous research study modified YOLOv8 to detect electrical pylons from satellite imagery, comparing it against baseline YOLOv8 but not against alternative architecture families. One study highlighted that while YOLOv8 provides a strong baseline for aerial power infrastructure detection, wide-area deployment via satellite remains constrained by the model's handling of complex shapes and small object sizes \citep{chi2024improved}. Another study states that both YOLO and Faster R-CNN have been applied to power infrastructure detection tasks, though they are rarely compared against each other on the same dataset, reinforcing the gap that our multi-architecture, single-dataset approach directly addresses \citep{faisal2025deep}.

Based on the review carried out, our study aims to fill a major gap in the literature: component-level identification, specifically, the detection of individual electrical sub-components. To do so, we are evaluating 16 model configurations: YOLOv8, YOLOv11, and YOLOv12, each in nano (n), small (s), medium (m), and large (l) variants, RF-DETR Base and Large, and Cascade R-CNN with 50-layer and 101-layer residual network (ResNet50 and ResNet101) backbones on a single dataset.

\clearpage
\section{Model Training and Substation Mapping}

In this section, we aim to develop and train models that can identify and differentiate between substation components. Our goal is to determine which model is the most accurate and precise, as well as to compare efficiencies in training time and inference speed. From there, we want to use the selected mapping model to make predictions on substations across the contiguous United States (CONUS), using imagery from the National Agriculture Imagery Program (NAIP).

For regional analysis, we use FERC Order 1000 (\citeyear{ferc2011order1000}) transmission planning regions, including Midcontinent Independent System Operator (MISO), Pennsylvania-New Jersey-Maryland interconnection (PJM), Southwest Power Pool (SPP), Electric Reliability Council of Texas (ERCOT), Florida Reliability Coordinating Council (FRCC), Southeastern Regional Transmission Planning (SERTP), South Carolina Regional Transmission Planning (SCRTP), New York Independent System Operator (NYISO), Independent System Operator New England (ISO-NE; ISONE in the figures), California Independent System Operator (CAISO), WestConnect Regional Transmission Planning (WestConnect), and NorthernGrid Regional Transmission Planning (NorthernGrid).

\begin{figure}[htbp]
\centering
\includegraphics[width=\textwidth,height=0.78\textheight,keepaspectratio]{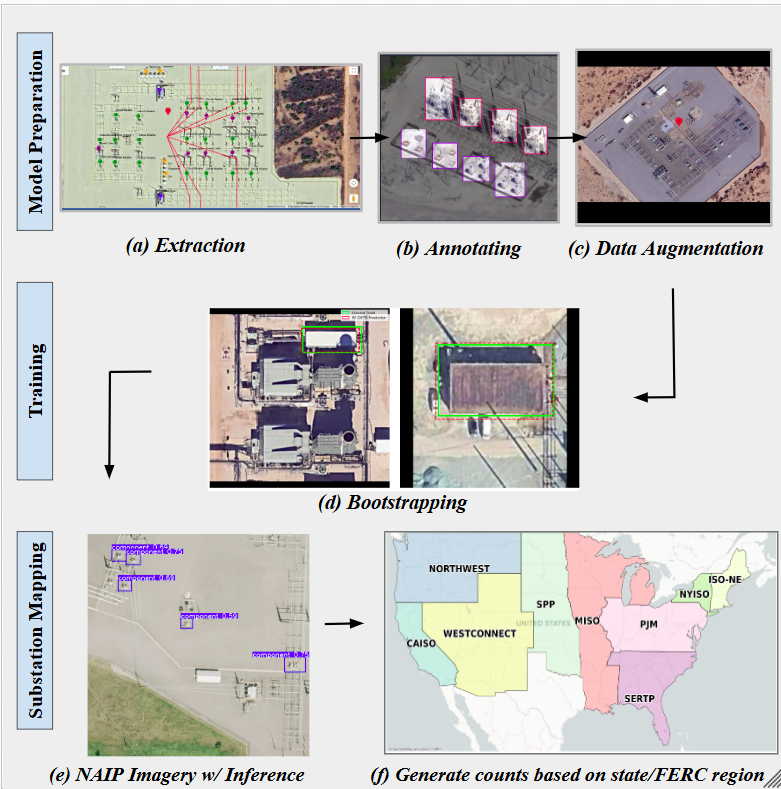}
\caption{This figure showcases each of the steps used in this section, from model preparation and data augmentation to running inference on substation imagery.}
\label{fig:workflow}
\end{figure}

We will go through each of the steps required for both model preparation and data augmentation, as shown in the figure above. This will lead us to create, label, and expand our dataset to allow for model training to occur.

\subsection{Model Preparation}

\subsubsection{Dataset Creation}

Our dataset will be built using open-source imagery of U.S. substations and transmission lines. Specifically, we will utilize a full-stack web-based geographic information system (webGIS) user interface, which leverages Google Earth's satellite and aerial imagery \citep{cmap}. Using this website, we will be able to extract images of electrical substations, which we can train our model on.

\subsubsection{Dataset Labeling}

From this application, we will collect 250 images for each component. As individual images of substations could contain more than one image class, some images were included in multiple class collections. Each image typically contains multiple substation components, which requires individual labeling. We will manually annotate each image by drawing precise boxes around each substation component present in the picture and assigning the appropriate class label, as shown in Figure (\ref{fig:workflow}b). We will consolidate these images into one dataset, and after removing any duplicate images across class-specific conditions, we will create a dataset of 1,389 images.

\subsubsection{Pre-Processing}

We will split our dataset of 1,389 images into training, validation, and testing data (via a 75-15-10 split), resulting in a total of 1,042 training, 210 validation, and 137 test images \citep{ghahremani2025detecting,egi2026realtime,vrsalovic2025system}. Before training, to help ensure consistency and enhance model performance, we will apply standardized preprocessing techniques to the dataset \citep{sapkota2025ultralytics,pottier2025modular}. This consists of resizing and auto-orienting the images. Several machine learning models, including YOLO, require images to have a fixed image size. To accommodate this, the images will be resized to 640x640 using letterbox padding, which will put black edges around the images to resize them while maintaining the original aspect ratio of the image, as shown in Figure (\ref{fig:workflow}c). Auto-orientation will also be applied to the images to correct any misalignments, ensuring that all the components appear in standardized positions \citep{huang2024deep,baig2025small,mishra2025automatic}.

\subsection{Model Training}

After creating a dataset, it is necessary to determine which models the dataset will be trained on. We will utilize various model architectures, including single-stage, two-stage, and transformer-based architectures, to determine the optimal model for this case. Specifically, we will select one family that uses each architecture: YOLO for a single-stage approach, Cascade for a two-stage approach, and RF-DETR for a transformer-based approach \citep{cheng2026review,carion2020endtoend}. Comparing these architectures will allow us to evaluate their differences in inference speed and precision.

\subsubsection{YOLO Models Overview}

YOLO is a single-stage object detection model with advanced data augmentation, a focal loss function, and a PyTorch-based architecture for optimization. Due to the speed at which YOLO models can be trained and YOLO's prevalence in object detection, we will utilize YOLO to showcase the effects of changing model speed and model generation. To demonstrate this, we will use YOLOv8, YOLOv11, and YOLOv12, each in nano/small/medium/large variants \citep{ramos2025comprehensive,pan2025intelligent,chaman2026benchmarking,shalash2026benchmarking,ali2026benchmarking}.

The YOLO family has been widely applied across many domains. In the electrical substation component area specifically, many studies have utilized the YOLO family to detect electrical substation components. For example, one study applied YOLOv8 to real-time aerial inspection of high-voltage power line components using UAV footage, reaching speeds of over 243 frames per second (fps) on a graphics processing unit (GPU) while accurately detecting insulator caps and line hardware \citep{bellou2024aerial}. Similarly, another study modified YOLOv8 to detect electrical pylons from optical satellite imagery, including a fast spatial pyramid pooling module with dynamic snake convolution and large selective kernels (DSLSK-SPPF), and an efficient multiscale convolution detection head (EMS-Head) to improve detection of complex shapes \citep{chi2024improved}. In insulator-specific work, another study proposed adaptive path fusion YOLO (APF-YOLO), an enhanced model with a corresponding fusion neck, improving detection of diverse insulator sizes in UAV imagery \citep{aitelhaj2025apfyolov8}. In addition to power infrastructure detection, YOLO models have been used for construction safety monitoring for personal protective equipment (PPE) and near-miss detection \citep{ali2026benchmarking} and in medical imaging for fracture detection in X-rays \citep{ferdi2025lightweight}.

The commonality regarding substation infrastructure studies in this subject is that they share a limitation of evaluating/comparing YOLO models in isolation or comparing only within the YOLO family. For example, one study utilized five YOLO generations on power equipment but did not include any outside detection models such as transformer-based or two-stage detectors \citep{he2024comprehensive}. By including YOLO versions 8, 11, and 12 with all variants, we can compare inference speed with precision and compare against Cascade models and RF-DETR.

\subsubsection{RF-DETR Overview}

RF-DETR was developed and released by Roboflow \citep{robinson2025rfdetr}. Most recently developed, the RF-DETR model is based on the DINO version 2 (DINOv2) backbone and is especially useful for detecting images in highly cluttered areas. RF-DETR is the sole model to achieve 60+ mean Average Precision when trained on the COCO dataset as well, highlighting not only its precision but its efficiency as well \citep{robinson2025rfdetr}. We will utilize two models in the RF-DETR family: RF-DETR and RF-DETR-large.

In a direct comparison with YOLOv12, one study found that RF-DETR outperformed YOLOv12x, YOLOv12l, and YOLOv12n in both single-class and multi-class detection of green fruit in an intricate environment, achieving a precision of 0.946 and an mAP@50 of 0.830 \citep{sapkota2025rfdetr}. Similarly, another study utilized RF-DETR for container damage detection \citep{kumar2025container}. In medical imaging, one study applied RF-DETR to mitotic figure detection, achieving an F1 score (the harmonic mean of precision and recall) of 0.789 with a recall of 0.839 and precision of 0.746 \citep{giedziun2025mitotic}.

We will include RF-DETR because its DINOv2 backbone is specifically designed for cluttered, multi-class images where objects are tight and close together, which is often the case in substation imagery. Additionally, its DINOv2 backbone has demonstrated strong performance on domain generalization benchmarks, outperforming prior real-time models on RF100-VL, a benchmark specifically designed to measure generalization across real-world detection tasks \citep{robinson2025rfdetr}. This makes it a strong candidate for detecting electrical substation imagery.

\subsubsection{Cascade Models Overview}

The Cascade R-CNN architecture uses a sequential chain of detection trained on progressively higher IoU thresholds. This ``cascade'' method utilizes the outputs of one stage as the input for the next, utilizing ResNet50 and ResNet101 backbones \citep{cai2018cascade}.

Within the electrical substation domain, Cascade R-CNN has proven to be highly effective for the detection of intricate or smaller components. For instance, one study implemented a Cascade R-CNN framework to detect defects in transmission line components, including vibration dampers and insulators, finding that the ResNet101 backbone achieved the highest detection precision \citep{zhao2023defect}. Similarly, another study developed an improved Cascade R-CNN model incorporating the shifted-window transformer version 2 (Swin-v2) to address scale variation in UAV images of transmission line components \citep{dong2023improved}. Beyond power infrastructure, Cascade R-CNN has been applied to food recognition benchmarks, where Cascade models outperformed other models, and to remote sensing instance segmentation tasks for large multi-channel aerial imagery \citep{carvalho2021instance}.

By training Cascade R-CNN with ResNet50 and ResNet101 alongside YOLO and RF-DETR models, we can directly compare the architectures and assess whether the progressive improvements translate into increased model precision.

\subsubsection{Model Implementation and Evaluation}

To train the selected models, the annotated dataset will be exported and converted into the YOLO PyTorch plain-text (TXT) format for our YOLO-based models and the COCO JavaScript Object Notation (JSON) format for Cascade and RF-DETR. All models will be implemented in PyTorch using a GPU-accelerated environment and trained for up to 50 epochs. Training performance and model checkpoints will be logged across iterations. Following training, detection accuracy will be evaluated across both individual component classes and full-model performance using mAP at an IoU threshold of 0.50 (mAP@50) and mAP averaged over IoU thresholds from 0.50 to 0.95 in increments of 0.05 (mAP@50:95). Average precision for an individual class at an IoU threshold of 0.50 is denoted AP50.

\subsection{Model Bootstrapping}

We recognize that one run of the model testing data does not accurately reflect model performance. Therefore, to effectively quantify the uncertainty associated with the modeling approach, we will perform bootstrapping utilizing the 91 test images of the models with the following per-class breakdown: alternate energy systems have 54 annotations, circuit breakers have 153 annotations, transformers have 61 annotations, and reactors have 58 annotations. We will resample the test images with replacement 1,000 times, calculating the mAP@50, mAP@50:95, and per-class AP50 values for each substation component category. This will produce a distribution of scores, from which we will generate a 95\% confidence interval (CI) of values, which will be included in our result graphs in Section 4.

\subsection{Substation Mapping}

To observe the effectiveness of our model, we will utilize YOLOv8-Small for mapping out the number of components in substations across the United States of America. To do this, we will use a pre-made dataset of substations found using open-source Earth mapping imagery and import a comma-separated values (CSV) file of their coordinates.

From there, we will use the U.S. Department of Agriculture (USDA) NAIP to extract images from each coordinate. A square of approximately 150 m x 150 m will be extracted, and RGB imagery will be extracted from the most recent available imagery from NAIP. NAIP provides nationwide orthophotography at approximately 0.6 m spatial resolution (the length and width each pixel covers), which makes it suitable for identifying large substation components while remaining realistically publicly accessible. The mapping dataset contains 11,085 images at 2048 pixels.

We will take these images and run inference on them using the mapping model. We aggregate the predicted components from these images by state using the Census 2018 1:500,000 boundaries and by FERC Order 1000 transmission planning region using the supplied planning-region polygons. Geographic summaries retain images assigned to both a state or the District of Columbia (DC) and a planning region. Two images (identifiers 832 and 6153), containing five retained detections, fall outside the planning-region polygons and are excluded from these summaries. State densities use Census land area, and planning-region densities use the area of the dissolved polygons using coordinate system 5070 in the European Petroleum Survey Group (EPSG) registry; both are reported per 100 km$^2$. 

\clearpage
\section{Results}

\subsection{Model Training Results}

In this section, we examined the results of the model training and determined a model to use for substation mapping.

To evaluate model performance, we used two object detection metrics: mAP@50 and mAP@50:95, or Mean Average Precision, measuring how accurately the model detected and localized objects across all the classes by averaging precision scores across recall levels. The @50 threshold means a detection is counted as correct if the predicted bounding box overlaps with the true bounding box by at least 50\%. mAP@50:95 is a stricter metric that averages performance across ten overlap thresholds ranging from 50\% to 95\% in 5\% increments. AP50 considers an IoU threshold of 0.50 between predicted and ground-truth bounding boxes, whereas mAP@50:95 averages across multiple IoU thresholds, rewarding detectors that achieve more accurate localization across varying levels of strictness \citep{tian2024performance}.

\begin{figure}[htbp]
\centering
\includegraphics[width=1\linewidth]{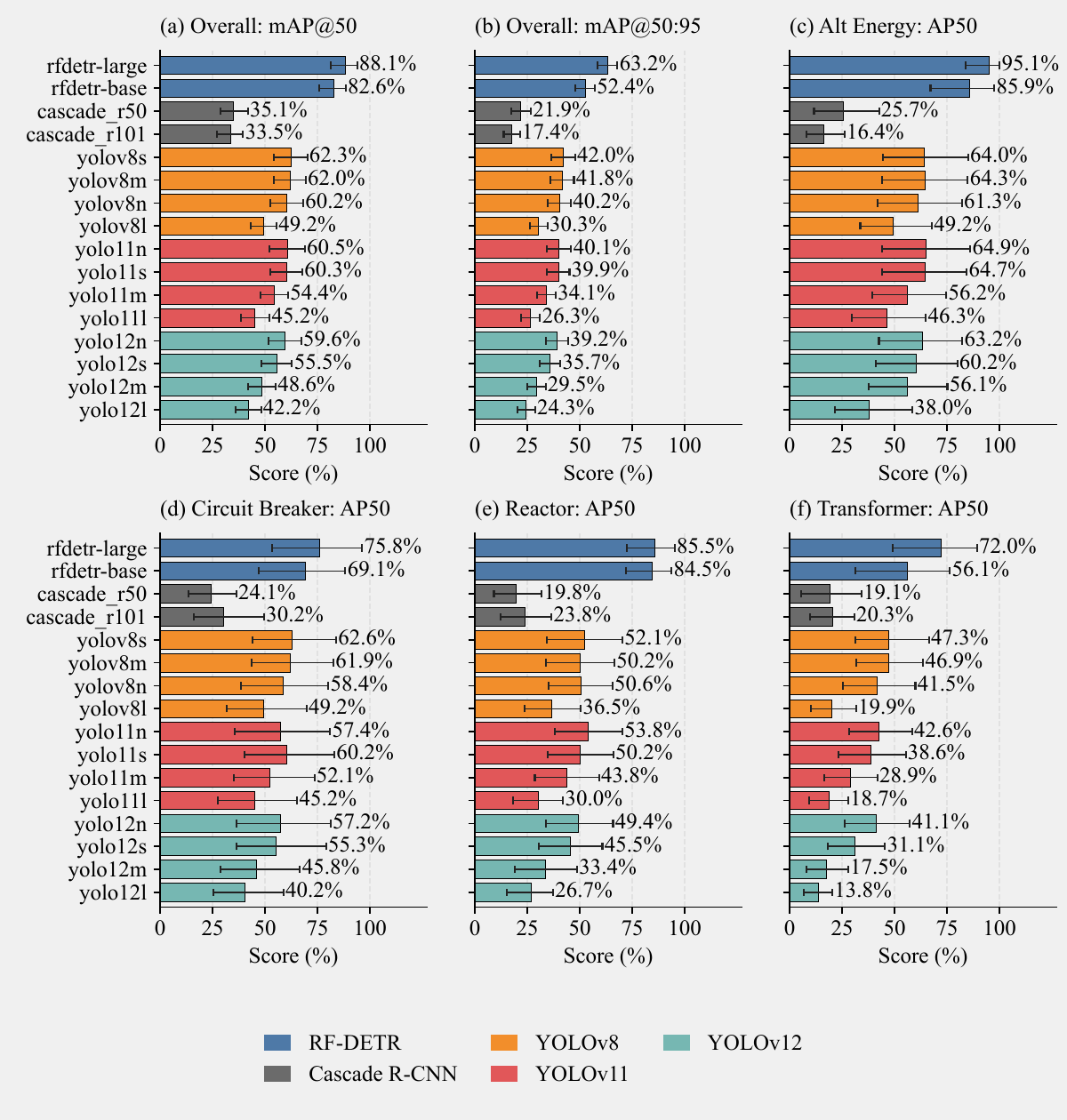}
\caption{The results of our model training. Panels (a) and (b) show overall model performance in terms of mAP@50 and mAP@50:95; panels (c)--(f) show per-class AP50. Alt Energy denotes alternate energy systems. Error bars represent 95\% confidence intervals calculated using 1,000 bootstrap resamples.}
\label{fig:benchmark}
\end{figure}

The figure above shows the mAP@50 and mAP@50:95 scores for all 16 models with their respective confidence intervals (CI). As shown in Figure (\ref{fig:benchmark}a), RF-DETR Large achieved the highest overall mAP@50 of 0.881, CI [0.814, 0.938], followed by RF-DETR Base at 0.826, CI [0.757, 0.884]. The RF-DETR family also achieved the highest overall mAP@50:95, with RF-DETR Large achieving an mAP@50:95 of 0.632, CI [0.585, 0.679], and RF-DETR Base achieving an mAP@50:95 of 0.5245, CI [0.478, 0.572]. RF-DETR's heightened performance compared to the other models is due to its heavily trained DINOv2 backbone and ability to perform well in complex and obstructed areas.

The next best-performing model family was YOLO. In Figure (\ref{fig:benchmark}), YOLO versions 8, 11, and 12 are shown, plotted for each model size (nano/small/medium/large). The best performing model was YOLOv8s, with an mAP@50 of 0.623, CI [0.541, 0.701] and mAP@50:95 of 0.420, CI [0.363, 0.477], whereas the worst performing model was YOLOv12l, with an mAP@50 of 0.422, CI [0.361, 0.481], and an mAP@50:95 of 0.243, CI [0.202, 0.286].

A trend emerged where, for the most part, smaller model architectures and older models performed better on this dataset. In YOLOv11 and YOLOv12, the nano-sized YOLO version had the highest mAP@50 (0.605, CI [0.523, 0.689] for YOLOv11, and 0.596, CI [0.517, 0.672] for YOLOv12) and mAP@50:95 values, followed by the small, medium, and large values. In YOLOv8, the small version performed the best, followed by the medium value, but the nano version was within 0.020 mAP@50 and 0.018 mAP@50:95 of the small value. This result can be attributed to overfitting. As our dataset is relatively smaller, larger models can overtrain on the data, resulting in a worse overall performance.

Additionally, we observed that older YOLO model architectures performed better than newer ones on this dataset. The average mAP@50 for YOLOv8 was 0.584, which was higher than YOLOv11's mAP@50 of 0.551 and YOLOv12's mAP@50 of 0.515. The average mAP@50:95 for YOLOv8 was 0.386, which was higher than YOLOv11's mAP@50:95 of 0.351 and YOLOv12's mAP@50:95 of 0.322. The reason older YOLO architectures perform better is that our dataset is relatively small and its domain is relatively unique, so the architectural complexity of newer YOLO models doesn't translate into improved performance, as consistent with prior studies \citep{sengun2025comparative}.

\subsection{Most Accurately Detected Components}

Based on our model training, we found that the most accurately detected component was alternate energy systems, with an average per-class AP50 of 0.570. Alternate systems are often the most visually distinct and have a repetitive rectangular and rack-like structure, allowing the models to detect them consistently regardless of model type. Circuit breakers had a lower average, with an average per-class AP50 of 0.528, whereas the least accurately detected were reactors, with an average AP50 of 0.460, and transformers with an average AP50 of 0.347. They are smaller and less visually distinctive, making the models harder to detect.

\subsection{Substation Mapping Results}

\begin{figure}[p]
\centering
\includegraphics[width=1\linewidth]{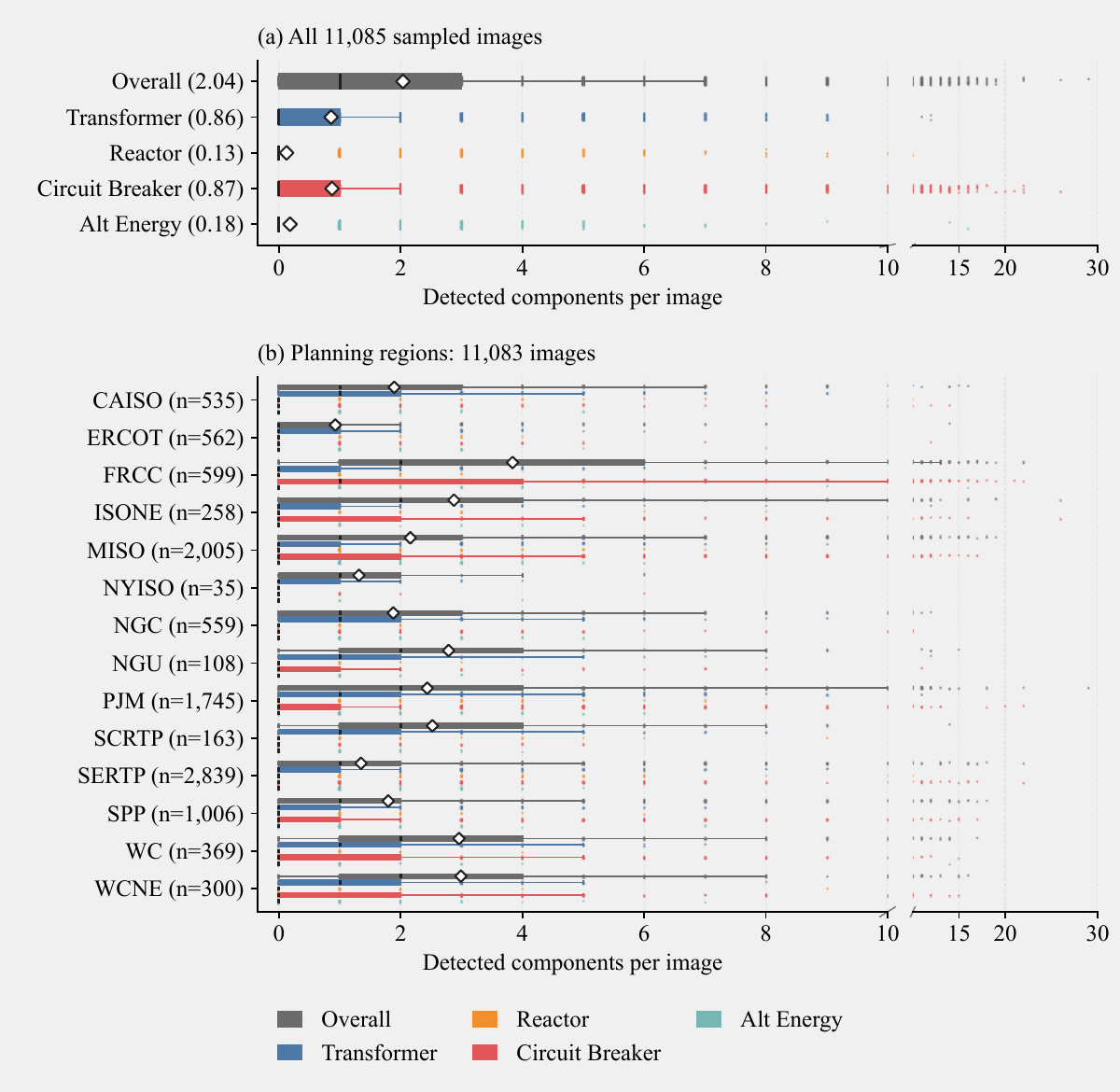}
\caption{Detected components per sampled image: (a) all 11,085 images, with means in parentheses; (b) the 11,083 geographically assigned images, with regional sample sizes. Boxes show interquartile ranges and medians; whiskers extend to observations within 1.5 interquartile ranges. Points represent images and diamonds indicate means. Zero-detection images are retained; Overall sums the four retained classes, excluding Control and Power Lines. The horizontal scale changes beyond 10 detections. NGC denotes NorthernGridConnected, NGU NorthernGridUnconnected, WCNE WestConnectNonEnrolled, and WC WestConnect.}
\label{fig:distributions}
\end{figure}

Overall, across the full dataset of 11,085 NAIP images analyzed, the model produced 22,596 detections, with an average of 2.04 detections per image (median of 1, interquartile range (IQR) from 0 to 3) within the four classes retained. As shown in Figure (\ref{fig:distributions}a), circuit breakers (averaging 0.87 detections/image) and transformers (0.86 detections/image) were detected the most frequently, whereas reactors (0.13 detections/image) and alternate energy systems (0.18 detections/image) were significantly lower. These distributions were skewed to the right; all four component classes had a median of zero detections, with a smaller number of images containing significantly more detections.

Detections also varied based on the FERC region. As shown in Figure (\ref{fig:distributions}b), overall patterns were consistent with those shown in Figure (\ref{fig:distributions}a), with transformers and circuit breakers having larger IQRs and overall distributions than reactors and alternate energy systems. The largest overall mean was within the FRCC region (3.84 detections/image) and the smallest overall mean was within the ERCOT region (0.93 detections/image). FRCC also had the largest overall IQR, from 1 to 6 detections/image, while ERCOT had the smallest IQR from 0 to 1 detections/image. These results provide image-level context for the subsequent overall state/FERC region-based geographic analyses.

\begin{figure}[htbp]
\centering
\includegraphics[width=\textwidth,height=0.78\textheight,keepaspectratio]{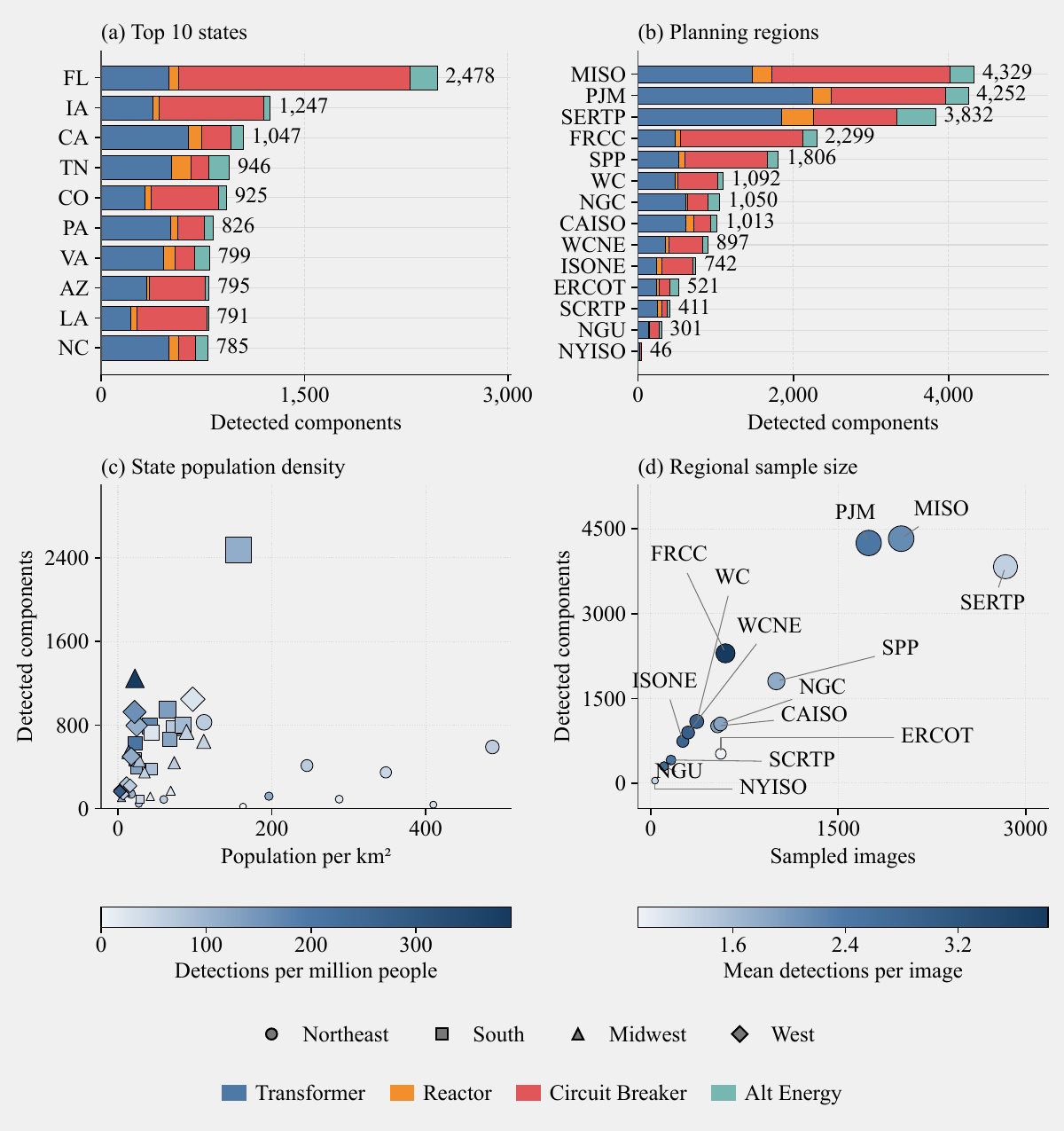}
\caption{Component detections by state (top ten; a) and FERC Order 1000 planning region (b), compared with state population density (c) and regional sampled image count (d). Panel (c) includes 47 sampled states with supplied population values; DC is excluded. Colour denotes detections per million people in (c) and mean detections per sampled image in (d). Marker size denotes total detections; shapes in (c) identify broad U.S. regions. Panel (d) describes sample coverage and does not estimate regional population density.}
\label{fig:counts}
\end{figure}

The state labels in Figure (\ref{fig:counts}a) denote Florida (FL), Iowa (IA), California (CA), Tennessee (TN), Colorado (CO), Pennsylvania (PA), Virginia (VA), Arizona (AZ), Louisiana (LA), and North Carolina (NC).

The figure above shows the number of components present based on state and FERC Order 1000 planning region, utilizing a stacked bar chart to describe the variation between the different components.

Figure (\ref{fig:counts}a) shows Florida's dominance in terms of the number of components measured in the state, with a total of 2,478 components. This is almost double the second-highest state, Iowa, with 1,247 components. Figure (\ref{fig:counts}b) shows that MISO (with 4,329 components), PJM (with 4,252 components), and SERTP (with 3,832 components) dominate component totals throughout FERC Order 1000 planning regions.

Figure (\ref{fig:counts}c) showcases a scatter plot that compares U.S. states based on their total number of infrastructure components (y-axis) against their population density (x-axis). The size of each data point is proportional to the number of how many substation components are present in that specific state. The color of each point indicates the components present per 1 million people. Different markers (circle, square, triangle, diamond) are used to categorize states into geographical regions (Northeast, South, Midwest, West)

Across the 47 sampled states and DC, component detections ranged from 17 (in DC) to 2,478 (in Florida), with a mean of approximately 471 detections and a median of 384 detections. Washington had no sampled images. The population-density comparison includes the 47 sampled states with supplied population values and excludes DC. This graph displayed only a weak positive association (Spearman's $\rho$ = 0.172, p = 0.247) between population density and substation component count, with no statistical significance. Florida represented an extreme in this graph; it had the highest number of components despite a population density of 157 people per square kilometer, which is less than a third of New Jersey's 487 people per square kilometer. Conversely, northeastern states such as New Jersey, Rhode Island, Massachusetts, Connecticut, and Maryland demonstrated an opposite pattern, due to their population densities of over 200 people/km2, but low component numbers (less than 600 each).

Figure (\ref{fig:counts}d) compares U.S. FERC Order 1000 planning regions based on their total number of detected components (y-axis) against the number of sampled substation images (x-axis). Similar to Figure (\ref{fig:counts}c), the size of each data point is proportional to the number of detections in that region, while the color of each point signifies the mean detections per sampled image.

The range of detections was larger for FERC regions, from MISO's 4,329 components to NYISO's 46 components. Across the 14 sampled regions, the mean number of components detected was approximately 1,614, and the median was 1,031.5. The regional scatter plot describes detections in relation to sampling coverage.

The state analysis did not show a statistically significant association between population density and component detections; it does not establish causation.

\begin{figure}[!t]
\centering
\includegraphics[width=0.95\linewidth,height=0.78\textheight,keepaspectratio]{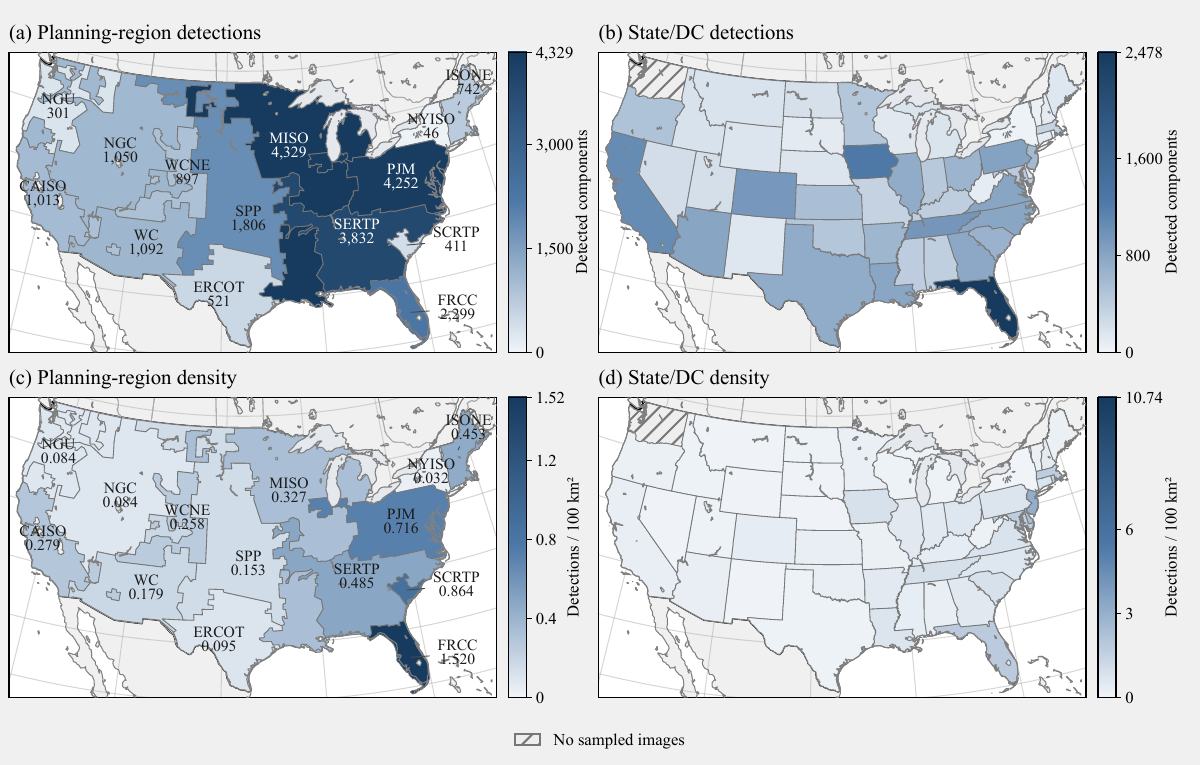}
\caption{Choropleth maps showing component detections by planning region (a) and state/DC (b), and detections per 100 km$^2$ by planning region (c) and state/DC (d). Both geographic summaries include 11,083 images and 22,591 detections. State densities use Census land area; regional densities use dissolved polygon area in EPSG:5070. Hatching indicates no sampled images.}
\label{fig:maps}
\end{figure}

All figures in this panel plot are choropleth maps, either describing the total component counts or component densities based on state or FERC region. Figure (\ref{fig:maps}a) shows the total number of detected components within each FERC Order 1000 planning region, matching the results shown in Figure (\ref{fig:counts}b), with MISO, PJM, and SERTP having the most detections out of the FERC regions. Figure (\ref{fig:maps}b) shows the total number of detected components within each U.S. state and DC, matching the state counts summarized in Figure (\ref{fig:counts}a), with Florida having the most detections. Figure (\ref{fig:maps}c) shows the density of detections per 100 square kilometers within each FERC Order 1000 planning region, where FRCC has the highest density (1.520 detections/100 km$^2$), followed by SCRTP (0.864 detections/100 km$^2$). Figure (\ref{fig:maps}d) shows the density of detections per 100 square kilometers within each state and DC, in which smaller northeastern states such as New Jersey (3.097 detections/100 km$^2$) and Delaware (2.358 detections/100 km$^2$) are higher than Florida (1.783 detections/100 km$^2$). DC has 10.736 detections/100 km$^2$.

\clearpage
\section{Discussion}

Here, we aim to go back and review the results of the model training and inference mapping in the context of the initial research questions.

\begin{enumerate}[
    label=\arabic*.,
    align=left,
    leftmargin=0pt,
    labelindent=0pt,
    itemindent=2em,
    labelwidth=1.5em,
    labelsep=0.5em
]
    
\item \textit{What is the best object detection model that can be utilized to identify and differentiate between substation components effectively?}

As shown in Figure (\ref{fig:benchmark}), RF-DETR Large achieved the highest mAP@50 among the sixteen models tested, reaching an mAP@50 of 0.881. Among the YOLO family, the YOLOv8-Small variant performed the best at 0.623, outperforming the other YOLOv11 and YOLOv12 variants.

\vspace{\baselineskip}

\item \textit{What components are most accurately detected and why?}

Across all 16 models, alternate energy systems were the most accurately detected class, reaching the highest AP50 of 0.951. Since alternate systems tend to be large and share repetitive rack-like structures, their appearances are clearer and easier for models to detect. Transformers and reactors were the most difficult classes, averaging 0.347 and 0.46, respectively. Especially since these components tend to be smaller, the models did not detect them as well. Circuit breakers fell in the middle range, performing moderately across all model families. Overall, we have observed that object size, their distinctiveness relative to their background, and their uniformity affect how well the models detect them.

Detection models, regardless of what type they are, usually require clear and well-defined features to draw accurate bounding boxes and assign them to the correct classes. Larger components like alternate energy systems occupy more pixels and have more consistent recognizable shapes across images, giving the models more to learn from. Smaller components like reactors and transformers are harder to distinguish due to their smaller size and the higher variability present in their dataset.

\vspace{\baselineskip}

\item \textit{Can we effectively count the total number of nationwide substation components in the United States and analyze them by both state and Federal Energy Regulatory Commission (FERC) Order 1000 planning regions?}

Using the mapping model, we obtained 22,591 component detections across the 11,083 sampled substation images retained in the geographic analysis within the United States.
\end{enumerate}
According to Figure (\ref{fig:counts}a), Florida led the nation with 2,478 total components, almost double the next closest state (Iowa with 1,247 components). Across these top-10 states, Circuit Breakers and Transformers made up the majority of the component mix, with Alternate Energy contributing smaller shares in states like Colorado and Tennessee. Furthermore, in Figure (\ref{fig:maps}b), we found that a small cluster of states, mostly in the Southeast and Central U.S., had the highest component counts. Figures (\ref{fig:counts}b) and (\ref{fig:maps}a) supported the notion that MISO, PJM, and SERTP had the highest component counts of 4,329, 4,252, and 3,832, respectively, which had almost double the number of components as the next FERC Order 1000 planning region cluster (FRCC). A third, much smaller group of FERC Order 1000 planning regions, NGC, CAISO, ERCOT, etc., has half of the previous; NYISO was the only outlier with 46 components.

Figure (\ref{fig:counts}c) compares states by total detections against population density, with color representing detections per 1 million people. Figure (\ref{fig:counts}d) compares FERC Order 1000 planning regions by total detections against sampled image count, with color representing mean detections per sampled image. While MISO and PJM had the highest totals, FRCC had the highest mean detections per sampled image. Figures (\ref{fig:maps}d) and (\ref{fig:maps}c) map density directly by state/DC and FERC Order 1000 planning region, respectively, without reference to population density.

Together, these results confirmed that our detection pipeline could identify individual components and analyze them by both state and FERC Order 1000 planning region.

\clearpage
\section{Conclusion}

In this paper, we evaluated the effectiveness and efficiency of 16 object detection model configurations spanning YOLOv8, YOLOv11, YOLOv12, RF-DETR, and Cascade for the automated mapping of substation components, including transformers, reactors, circuit breakers, and alternate energy infrastructure. We performed a literature review to identify past/current gaps in substation detection methodology. Our study aimed to address these gaps by using the same dataset to train all three model families and to compare the models. Our dataset included images taken from open-source imagery of U.S. substation equipment.

All in all, the RF-DETR Large model achieved the highest mean average precision (mAP) of 0.881, with a CI of 0.814 to 0.938, while YOLOv8-Small performed best among the YOLO family, with the highest overall detection accuracy of 0.623 with a CI of 0.541 to 0.701. While YOLO variants offered significantly faster inference speeds, RF-DETR Large achieved the highest precision and well-rounded performance across all substation components. Our use of the mapping model on national-scale imagery demonstrated their practical, real-world utility, rather than serving as a baseline comparison. Ultimately, this research not only offers a clear, evidence-based foundation for selecting detection models suited to specific precision and speed constraints but also a framework for automated substation infrastructure counts for the future.

\clearpage
\section*{Acknowledgments}

Dennies Kiprono Bor, Dante Groccia, and Edward J. Oughton gratefully acknowledge that this material is based upon work supported by the National Center for Atmospheric Research (NCAR), which is a major facility sponsored by the U.S. National Science Foundation (NSF) under Cooperative Agreement No. 1852977. The paper was produced via a project funded through supplemental funding from NSF NCAR Faculty Innovator Program under the same Cooperative Agreement. Edward J. Oughton also acknowledges research support via the Integrated Hazard Institute under National Aeronautics and Space Administration (NASA) Cooperative Agreement 80NSSC25M0077.

\clearpage
\section*{Open Research}

The supplementary material for this project, including the code and data utilized for model preparation, inference mapping, geographic aggregation, and figure creation, is included within GitHub, of which releases are synced to Zenodo. The digital object identifier (DOI) for this paper's Zenodo is \url{https://doi.org/10.5281/zenodo.22239093}.

\clearpage
\section*{Conflict of Interest}

No conflicts of interest.

\clearpage
\section*{Author Contributions}

Namish Bansal: Writing, Graph Creation, Data Collection, Model Training; Haley Mody: Writing, Graph Creation, Data Collection, Model Training; Dennies Kiprono Bor: Supervision, Writing-reviewing; Dante Groccia: Supervision, Writing-reviewing; Edward J. Oughton: Supervision, Writing-reviewing.

\clearpage
\bibliographystyle{abbrvnat}
\bibliography{references}

@misc{abdelfattah2020ttpla,
  author = {Abdelfattah, Rabab and Wang, Xiaofeng and Wang, Song},
  title = {{TTPLA: An Aerial-Image Dataset for Detection and Segmentation of Transmission Towers and Power Lines}},
  year = {2020},
  doi = {10.48550/arxiv.2010.10032},
  eprint = {2010.10032v1},
  archivePrefix = {arXiv},
  howpublished = {arXiv preprint 2010.10032v1},
  url = {https://arxiv.org/abs/2010.10032v1},
}

@article{aitelhaj2025apfyolov8,
  author = {Aitelhaj, Rita and Benelmostafa, Badr-Eddine and Medromi, Hicham},
  title = {{APF-YOLOV8: Enhancing Multiscale Detection and Intra-Class Variance Handling for UAV-Based Insulator Power Line Inspections}},
  year = {2025},
  doi = {10.12688/f1000research.160650.2},
  journal = {F1000Research},
  volume = {14},
  pages = {141},
}

@article{ali2026benchmarking,
  author = {Ali, Shahid and Alzarrad, Ammar and Chowdhury, Sudipta and Narman, Husnu S.},
  title = {{Benchmarking YOLOv8-YOLOv12 models for image-space near-miss detection in construction safety}},
  year = {2026},
  doi = {10.3389/fbuil.2026.1827664},
  journal = {Frontiers in Built Environment},
  volume = {12},
  eid = {1827664},
  pages = {1827664},
}

@misc{baig2025small,
  author = {Baig, Danish Zia and Kamal, Mohsin and Ullah, Zahid},
  title = {{Small Dents, Big Impact: A Dataset and Deep Learning Approach for Vehicle Dent Detection}},
  year = {2025},
  doi = {10.48550/arxiv.2508.15431},
  eprint = {2508.15431v2},
  archivePrefix = {arXiv},
  howpublished = {arXiv preprint 2508.15431v2},
  url = {https://arxiv.org/abs/2508.15431v2},
}

@article{bao2022defect,
  author = {Bao, Wenxia and Du, Xiang and Wang, Nian and Yuan, Mu and Yang, Xianjun},
  title = {{A Defect Detection Method Based on BC-YOLO for Transmission Line Components in UAV Remote Sensing Images}},
  year = {2022},
  doi = {10.3390/rs14205176},
  journal = {Remote Sensing},
  volume = {14},
  number = {20},
  pages = {5176},
}

@article{bellou2024aerial,
  author = {Bellou, Elisavet and Pisica, Ioana and Banitsas, Konstantinos},
  title = {{Aerial Inspection of High-Voltage Power Lines Using YOLOv8 Real-Time Object Detector}},
  year = {2024},
  doi = {10.3390/en17112535},
  journal = {Energies},
  volume = {17},
  number = {11},
  pages = {2535},
}

@article{cmap,
  author = {Bor, Dennies K. and Oughton, Edward J. and Peters, Evan A. and Rivera, Noah and Gaunt, C. Trevor and Weigel, Robert S. and Wiltberger, Michael},
  title = {{Systematic Component-Level Characterization of Electricity Transmission Infrastructure for Space Weather Risk Assessment}},
  journal = {Space Weather},
  year = {2026},
  volume = {24},
  number = {9},
  pages = {e2026SW005166},
  doi = {10.1029/2026SW005166},
}

@inproceedings{cai2018cascade,
  author = {Cai, Zhaowei and Vasconcelos, Nuno},
  title = {{Cascade R-CNN: Delving Into High Quality Object Detection}},
  year = {2018},
  doi = {10.1109/cvpr.2018.00644},
  booktitle = {2018 IEEE/CVF Conference on Computer Vision and Pattern Recognition},
  pages = {6154--6162},
}

@misc{carion2020endtoend,
  author = {Carion, Nicolas and Massa, Francisco and Synnaeve, Gabriel and Usunier, Nicolas and Kirillov, Alexander and Zagoruyko, Sergey},
  title = {{End-to-End Object Detection with Transformers}},
  year = {2020},
  doi = {10.48550/arxiv.2005.12872},
  eprint = {2005.12872v3},
  archivePrefix = {arXiv},
  howpublished = {arXiv preprint 2005.12872v3},
  url = {https://arxiv.org/abs/2005.12872v3},
}

@article{carvalho2021instance,
  author = {Carvalho, Osmar Luiz Ferreira de and de Carvalho J{\'{u}}nior, Osmar Ab{\'{i}}lio and Albuquerque, Anesmar Olino de and Bem, Pablo Pozzobon de and Silva, Cristiano Rosa and Ferreira, Pedro Henrique Guimar{\~{a}}es and Moura, Rebeca dos Santos de and Gomes, Roberto Arnaldo Trancoso and Guimar{\~{a}}es, Renato Fontes and Borges, D{\'{i}}bio Leandro},
  title = {{Instance Segmentation for Large, Multi-Channel Remote Sensing Imagery Using Mask-RCNN and a Mosaicking Approach}},
  year = {2021},
  doi = {10.3390/rs13010039},
  journal = {Remote Sensing},
  volume = {13},
  number = {1},
  pages = {39},
}

@article{chaman2026benchmarking,
  author = {Chaman, Mohammed and Maliki, Anas El and Dahou, Hamad and Hadjoudja, Abdelkader},
  title = {{Benchmarking YOLO-based deep learning models for real-time object detection in hybrid ADAS and intelligent transportation systems}},
  year = {2026},
  doi = {10.1016/j.rineng.2025.108942},
  journal = {Results in Engineering},
  volume = {29},
  pages = {108942},
  eid = {108942},
}

@article{chandra2025novel,
  author = {Chandra, Naveen and Vaidya, Himadri and Satyam, Neelima and Tang, Xiaochuan and Singh, Saurabh and Meena, Sansar Raj},
  title = {{A Novel Multi-Layer Attention Boosted YOLOv10 Network for Landslide Mapping Using Remote Sensing Data}},
  year = {2025},
  doi = {10.1111/tgis.70023},
  journal = {Transactions in GIS},
  volume = {29},
  number = {2},
  eid = {e70023},
  pages = {e70023},
}

@article{chen2023insuyolo,
  author = {Chen, Yifu and Liu, Hongye and Chen, Jiahao and Hu, Jianhong and Zheng, Enhui},
  title = {{Insu-YOLO: An Insulator Defect Detection Algorithm Based on Multiscale Feature Fusion}},
  year = {2023},
  doi = {10.3390/electronics12153210},
  journal = {Electronics},
  volume = {12},
  number = {15},
  pages = {3210},
}

@misc{chen2024lwdetr,
  author = {Chen, Qiang and Su, Xiangbo and Zhang, Xinyu and Wang, Jian and Chen, Jiahui and Shen, Yunpeng and Han, Chuchu and Chen, Ziliang and Xu, Weixiang and Li, Fanrong and Zhang, Shan and Yao, Kun and Ding, Errui and Zhang, Gang and Wang, Jingdong},
  title = {{LW-DETR: A Transformer Replacement to YOLO for Real-Time Detection}},
  year = {2024},
  doi = {10.48550/arxiv.2406.03459},
  eprint = {2406.03459v1},
  archivePrefix = {arXiv},
  howpublished = {arXiv preprint 2406.03459v1},
  url = {https://arxiv.org/abs/2406.03459v1},
}

@article{cheng2026review,
  author = {Cheng, Bei and Deng, Quanli and Gan, WenJie and Li, Yunlong and Wu, Changneng and Wang, Qingwang},
  title = {{A Review of Remote Sensing Image Object Detection: Advances and Challenges in Data, Techniques, and Applications}},
  year = {2026},
  doi = {10.1109/jstars.2026.3671199},
  journal = {IEEE Journal of Selected Topics in Applied Earth Observations and Remote Sensing},
  volume = {19},
  pages = {10438--10472},
}

@article{chi2024improved,
  author = {Chi, Xin and Sun, Yu and Zhao, Yingjun and Lu, Donghua and Gao, Yan and Zhang, Yiting},
  title = {{An Improved YOLOv8 Network for Detecting Electric Pylons Based on Optical Satellite Image}},
  year = {2024},
  doi = {10.3390/s24124012},
  journal = {Sensors},
  volume = {24},
  number = {12},
  pages = {4012},
}

@article{dong2023improved,
  author = {Dong, Chao and Zhang, Ke and Xie, Zhiyuan and Shi, Chaojun},
  title = {{An improved cascade RCNN detection method for key components and defects of transmission lines}},
  year = {2023},
  doi = {10.1049/gtd2.12948},
  journal = {IET Generation, Transmission \& Distribution},
  volume = {17},
  number = {19},
  pages = {4277--4292},
}

@article{egi2026realtime,
  author = {Egi, Yunus and Oter, T{\"{u}}lay and Hajyzadeh, Mortaza and Catak, Muammer},
  title = {{Real-Time Callus Instance Segmentation in Plant Tissue Culture Using Successive Generations of YOLO Architectures}},
  year = {2026},
  doi = {10.3390/plants15010047},
  journal = {Plants},
  volume = {15},
  number = {1},
  pages = {47},
}

@article{erenoglu2024power,
  author = {Ereno{\u{g}}lu, Ay{\c{s}}e K{\"{u}}bra and Sengor, Ibrahim and Erdin{\c{c}}, Ozan},
  title = {{Power System Resiliency: A Comprehensive Overview from Implementation Aspects and Innovative Concepts}},
  year = {2024},
  doi = {10.1016/j.nexus.2024.100311},
  journal = {Energy Nexus},
  volume = {15},
  pages = {100311},
  eid = {100311},
}

@article{faisal2025deep,
  author = {Faisal, Md. Ahasan Atick and Mecheter, Imene and Qiblawey, Yazan and Fernandez, Javier Hernandez and Chowdhury, Muhammad E.H. and Kiranyaz, Serkan},
  title = {{Deep learning in automated power line inspection: A review}},
  year = {2025},
  doi = {10.1016/j.apenergy.2025.125507},
  journal = {Applied Energy},
  volume = {385},
  pages = {125507},
  eid = {125507},
}

@techreport{ferc2011order1000,
  author = {{Federal Energy Regulatory Commission}},
  title = {{Transmission Planning and Cost Allocation by Transmission Owning and Operating Public Utilities}},
  institution = {Federal Energy Regulatory Commission},
  type = {Final Rule},
  number = {Order No. 1000; Docket RM10-23-000},
  year = {2011},
  url = {https://www.ferc.gov/sites/default/files/2020-06/Order-1000.pdf},
}

@misc{ferdi2025lightweight,
  author = {Ferdi, Abdesselam},
  title = {{Lightweight G-YOLOv11: Advancing Efficient Fracture Detection in Pediatric Wrist X-rays}},
  year = {2025},
  doi = {10.48550/arxiv.2501.00647},
  eprint = {2501.00647v1},
  archivePrefix = {arXiv},
  howpublished = {arXiv preprint 2501.00647v1},
  url = {https://arxiv.org/abs/2501.00647v1},
}

@article{ghahremani2025detecting,
  author = {Ghahremani, Ali and Adams, Scott D. and Norton, Michael and Khoo, Sui Yang and Kouzani, Abbas Z.},
  title = {{Detecting Defects in Solar Panels Using the YOLO v10 and v11 Algorithms}},
  year = {2025},
  doi = {10.3390/electronics14020344},
  journal = {Electronics},
  volume = {14},
  number = {2},
  pages = {344},
}

@misc{giedziun2025mitotic,
  author = {Giedziun, Piotr and So{\l}tysik, Jan and G{\'{o}}rczany, Mateusz and Ropiak, Norbert and Przymus, Marcin and Krajewski, Piotr and Kwiecie{\'{n}}, Jaros{\l}aw and Bartczak, Artur and Wasiak, Izabela and Maniewski, Mateusz},
  title = {{RF-DETR for Robust Mitotic Figure Detection: A MIDOG 2025 Track 1 Approach}},
  year = {2025},
  doi = {10.48550/arxiv.2509.02599},
  eprint = {2509.02599v1},
  archivePrefix = {arXiv},
  howpublished = {arXiv preprint 2509.02599v1},
  url = {https://arxiv.org/abs/2509.02599v1},
}

@article{han2020electrical,
  author = {Han, Sheng and Yang, Fan and Yang, Gang and Gao, Bing and Zhang, Na and Wang, Dawei},
  title = {{Electrical equipment identification in infrared images based on ROI-selected CNN method}},
  year = {2020},
  doi = {10.1016/j.epsr.2020.106534},
  journal = {Electric Power Systems Research},
  volume = {188},
  pages = {106534},
  eid = {106534},
}

@misc{he2024comprehensive,
  author = {He, Zijian and Wang, Kang and Fang, Tian and Su, Lei and Chen, Rui and Fei, Xihong},
  title = {{Comprehensive Performance Evaluation of YOLOv11, YOLOv10, YOLOv9, YOLOv8 and YOLOv5 on Object Detection of Power Equipment}},
  year = {2024},
  doi = {10.48550/arxiv.2411.18871},
  eprint = {2411.18871v1},
  archivePrefix = {arXiv},
  howpublished = {arXiv preprint 2411.18871v1},
  url = {https://arxiv.org/abs/2411.18871v1},
}

@article{hu2025uav,
  author = {Hu, Cong and Lv, Lingfeng and Zhou, Tian},
  title = {{UAV inspection insulator defect detection method based on dynamic adaptation improved YOLOv8}},
  year = {2025},
  doi = {10.1007/s11554-025-01660-8},
  journal = {Journal of Real-Time Image Processing},
  volume = {22},
  number = {2},
  eid = {74},
  pages = {74},
}

@misc{huang2024deep,
  author = {Huang, Sida and Wu, Yuanting and Nguyen, Dinh Hoa},
  title = {{Deep Learning Based Solar Cell Recognition for Optical Wireless Power Transfer}},
  year = {2024},
  doi = {10.48550/arxiv.2410.14096},
  eprint = {2410.14096v1},
  archivePrefix = {arXiv},
  howpublished = {arXiv preprint 2410.14096v1},
  url = {https://arxiv.org/abs/2410.14096v1},
}

@article{hussain2023yolov1,
  author = {Hussain, Muhammad},
  title = {{YOLO-v1 to YOLO-v8, the Rise of YOLO and Its Complementary Nature toward Digital Manufacturing and Industrial Defect Detection}},
  year = {2023},
  doi = {10.3390/machines11070677},
  journal = {Machines},
  volume = {11},
  number = {7},
  pages = {677},
}

@article{imdadullah2021electric,
  author = {{Imdadullah} and Alamri, Basem and Hossain, Md. Alamgir and Asghar, M. S. Jamil},
  title = {{Electric Power Network Interconnection: A Review on Current Status, Future Prospects and Research Direction}},
  year = {2021},
  doi = {10.3390/electronics10172179},
  journal = {Electronics},
  volume = {10},
  number = {17},
  pages = {2179},
}

@article{jia2024defect,
  author = {Jia, Yanfei and Chen, Guangda and Zhao, Liquan},
  title = {{Defect detection of photovoltaic modules based on improved VarifocalNet}},
  year = {2024},
  doi = {10.1038/s41598-024-66234-3},
  journal = {Scientific Reports},
  volume = {14},
  number = {1},
  eid = {15170},
  pages = {15170},
}

@article{kalt2021global,
  author = {Kalt, Gerald and Thunshirn, Philipp and Haberl, Helmut},
  title = {{A global inventory of electricity infrastructures from 1980 to 2017: Country-level data on power plants, grids and transformers}},
  year = {2021},
  doi = {10.1016/j.dib.2021.107351},
  journal = {Data in Brief},
  volume = {38},
  pages = {107351},
  eid = {107351},
}

@article{kaur2024developments,
  author = {Kaur, Gaganpreet and Afaq, Yasir},
  title = {{Developments in deep learning for change detection in remote sensing: A review}},
  year = {2024},
  doi = {10.1111/tgis.13133},
  journal = {Transactions in GIS},
  volume = {28},
  number = {2},
  pages = {223--257},
}

@article{kaya2025downscaling,
  author = {Kaya, Asir Y{\"{u}}ksel and Sajjad, Muhammad and Dedekorkut-Howes, Aysin},
  title = {{Downscaling Urban Disaster Resilience: Toward Building-Scale Geo-Information Modeling and Machine Learning-Assisted Resilience Knowledge Systems}},
  year = {2025},
  doi = {10.1111/tgis.70153},
  journal = {Transactions in GIS},
  volume = {29},
  number = {8},
  eid = {e70153},
  pages = {e70153},
}

@article{khalili2024sodyolov8,
  author = {Khalili, Boshra and Smyth, Andrew W.},
  title = {{SOD-YOLOv8---Enhancing YOLOv8 for Small Object Detection in Aerial Imagery and Traffic Scenes}},
  year = {2024},
  doi = {10.3390/s24196209},
  journal = {Sensors},
  volume = {24},
  number = {19},
  pages = {6209},
}

@misc{kumar2025container,
  author = {Kumar, Subhadip},
  title = {{Container damage detection using advanced computer vision model Yolov12 vs Yolov11 vs RF-DETR A comparative analysis}},
  year = {2025},
  doi = {10.48550/arxiv.2506.22517},
  eprint = {2506.22517v1},
  archivePrefix = {arXiv},
  howpublished = {arXiv preprint 2506.22517v1},
  url = {https://arxiv.org/abs/2506.22517v1},
}

@article{kuznetsova2020open,
  author = {Kuznetsova, Alina and Rom, Hassan and Alldrin, Neil and Uijlings, Jasper and Krasin, Ivan and Pont-Tuset, Jordi and Kamali, Shahab and Popov, Stefan and Malloci, Matteo and Kolesnikov, Alexander and Duerig, Tom and Ferrari, Vittorio},
  title = {{The Open Images Dataset V4}},
  year = {2020},
  doi = {10.1007/s11263-020-01316-z},
  journal = {International Journal of Computer Vision},
  volume = {128},
  number = {7},
  pages = {1956--1981},
}

@article{lai2019automatic,
  author = {Lai, Qiupin and Yang, Jun and Tan, Bendong and Wang, Liang and Fu, Siyao and Han, Liwei},
  title = {{An Automatic Recognition and Defect Diagnosis Model of Transmission Line Insulator Based on YOLOv2 Network}},
  journal = {Electric Power},
  year = {2019},
  volume = {52},
  number = {7},
  pages = {31--39},
  doi = {10.11930/j.issn.1004-9649.201806102},
}

@misc{lavado2025enhancing,
  author = {Lavado, Diogo and Santos, Ricardo and Coelho, Andre and Santos, Joao and Micheletti, Alessandra and Soares, Claudia},
  title = {{Enhancing Power Grid Inspections with Machine Learning}},
  year = {2025},
  doi = {10.48550/arxiv.2502.13037},
  eprint = {2502.13037v1},
  archivePrefix = {arXiv},
  howpublished = {arXiv preprint 2502.13037v1},
  url = {https://arxiv.org/abs/2502.13037v1},
}

@article{lekidis2022electricity,
  author = {Lekidis, Alexios and Anastasiadis, Anestis G. and Vokas, Georgios A.},
  title = {{Electricity infrastructure inspection using AI and edge platform-based UAVs}},
  year = {2022},
  doi = {10.1016/j.egyr.2022.07.115},
  journal = {Energy Reports},
  volume = {8},
  pages = {1394--1411},
}

@article{li2025enhanced,
  author = {Li, Zhuyun and Qin, Qiutong and Yang, Yingyi and Mai, Xiaoming and Ieiri, Yuya and Yoshie, Osamu},
  title = {{An enhanced substation equipment detection method based on distributed federated learning}},
  year = {2025},
  doi = {10.1016/j.ijepes.2025.110547},
  journal = {International Journal of Electrical Power \& Energy Systems},
  volume = {166},
  pages = {110547},
  eid = {110547},
}

@article{li2026frnet,
  author = {Li, Tianyu and Zhu, Changsheng and Wang, Yongxin and Li, Menghan and Cao, Hang and Li, Jingjie and Yuan, Peiwen and Bai, Hongwei},
  title = {{FRNet: A lightweight feature reconfiguration network for infrared detection of substation equipment and instruments}},
  year = {2026},
  doi = {10.1016/j.dsp.2025.105513},
  journal = {Digital Signal Processing},
  volume = {168},
  pages = {105513},
  eid = {105513},
}

@article{liu2020research,
  author = {Liu, Yunpeng and Ji, Xinxin and Pei, Shaotong and Ma, Ziru and Zhang, Gonghao and Lin, Ying and Chen, Yufeng},
  title = {{Research on automatic location and recognition of insulators in substation based on YOLOv3}},
  year = {2020},
  doi = {10.1049/hve.2019.0091},
  journal = {High Voltage},
  volume = {5},
  number = {1},
  pages = {62--68},
}

@article{liu2026accurate,
  author = {Liu, Xiaolin and Quan, Wei and Liu, Xiaohong and Wang, Xiaoyu and Hong, Peiyuan and Xu, Xiaoqian},
  title = {{An accurate detection and segmentation model of traction substation equipment for high-speed railway based on cascade query embedding}},
  year = {2026},
  doi = {10.1080/23248378.2025.2495200},
  journal = {International Journal of Rail Transportation},
  volume = {14},
  number = {1},
  pages = {199--219},
}

@article{mattern2025comparison,
  author = {Mattern, Alexander and Gerdes, Henrik and Grunert, Dennis and Schmitt, Robert H.},
  title = {{A comparison of transformer and CNN-based object detection models for surface defects on Li-Ion Battery Electrodes}},
  year = {2025},
  doi = {10.1016/j.est.2024.114378},
  journal = {Journal of Energy Storage},
  volume = {105},
  pages = {114378},
  eid = {114378},
}

@article{meyur2022ensembles,
  author = {Meyur, Rounak and Vullikanti, Anil and Swarup, Samarth and Mortveit, Henning S. and Centeno, Virgilio and Phadke, Arun and Poor, H. Vincent and Marathe, Madhav V.},
  title = {{Ensembles of realistic power distribution networks}},
  year = {2022},
  doi = {10.1073/pnas.2205772119},
  journal = {Proceedings of the National Academy of Sciences},
  volume = {119},
  number = {42},
  eid = {e2205772119},
  pages = {e2205772119},
}

@article{miao2019insulator,
  author = {Miao, Xiren and Liu, Xinyu and Chen, Jing and Zhuang, Shengbin and Fan, Jianwei and Jiang, Hao},
  title = {{Insulator Detection in Aerial Images for Transmission Line Inspection Using Single Shot Multibox Detector}},
  year = {2019},
  doi = {10.1109/access.2019.2891123},
  journal = {IEEE Access},
  volume = {7},
  pages = {9945--9956},
}

@article{mishra2025automatic,
  author = {Mishra, Gangeshwar and Tanwar, Rohit and Gupta, Prinima},
  title = {{Automatic Target Recognition using Unmanned Aerial Vehicle Images with Proposed YOLOv8-SR and Enhanced Deep Super-Resolution Network}},
  year = {2025},
  doi = {10.35882/jeeemi.v7i4.888},
  journal = {Journal of Electronics, Electromedical Engineering, and Medical Informatics},
  volume = {7},
  number = {4},
  pages = {1240--1258},
}

@article{nguyen2019intelligent,
  author = {Nguyen, Van Nhan and Jenssen, Robert and Roverso, Davide},
  title = {{Intelligent Monitoring and Inspection of Power Line Components Powered by UAVs and Deep Learning}},
  year = {2019},
  doi = {10.1109/jpets.2018.2881429},
  journal = {IEEE Power and Energy Technology Systems Journal},
  volume = {6},
  number = {1},
  pages = {11--21},
}

@article{ni2024smallobject,
  author = {Ni, Jianjun and Zhu, Shengjie and Tang, Guangyi and Ke, Chunyan and Wang, Tingting},
  title = {{A Small-Object Detection Model Based on Improved YOLOv8s for UAV Image Scenarios}},
  year = {2024},
  doi = {10.3390/rs16132465},
  journal = {Remote Sensing},
  volume = {16},
  number = {13},
  pages = {2465},
}

@article{nie2024lightweight,
  author = {Nie, Haijiao and Pang, Huanli and Ma, Mingyang and Zheng, Ruikai},
  title = {{A Lightweight Remote Sensing Small Target Image Detection Algorithm Based on Improved YOLOv8}},
  year = {2024},
  doi = {10.3390/s24092952},
  journal = {Sensors},
  volume = {24},
  number = {9},
  pages = {2952},
}

@article{ohanu2024comprehensive,
  author = {Ohanu, Chibuike Peter and Rufai, Salihu Ahmed and Oluchi, Ugbe Christiana},
  title = {{A comprehensive review of recent developments in smart grid through renewable energy resources integration}},
  year = {2024},
  doi = {10.1016/j.heliyon.2024.e25705},
  journal = {Heliyon},
  volume = {10},
  number = {3},
  pages = {e25705},
  eid = {e25705},
}

@misc{oquab2023dinov2,
  author = {Oquab, Maxime and Darcet, Timoth{\'{e}}e and Moutakanni, Th{\'{e}}o and Vo, Huy and Szafraniec, Marc and Khalidov, Vasil and Fernandez, Pierre and Haziza, Daniel and Massa, Francisco and El-Nouby, Alaaeldin and Assran, Mahmoud and Ballas, Nicolas and Galuba, Wojciech and Howes, Russell and Huang, Po-Yao and Li, Shang-Wen and Misra, Ishan and Rabbat, Michael and Sharma, Vasu and Synnaeve, Gabriel and Xu, Hu and Jegou, Herv{\'{e}} and Mairal, Julien and Labatut, Patrick and Joulin, Armand and Bojanowski, Piotr},
  title = {{DINOv2: Learning Robust Visual Features without Supervision}},
  year = {2023},
  doi = {10.48550/arxiv.2304.07193},
  eprint = {2304.07193v2},
  archivePrefix = {arXiv},
  howpublished = {arXiv preprint 2304.07193v2},
  url = {https://arxiv.org/abs/2304.07193v2},
}

@article{padilla2021comparative,
  author = {Padilla, Rafael and Passos, Wesley L. and Dias, Thadeu L. B. and Netto, Sergio L. and da Silva, Eduardo A. B.},
  title = {{A Comparative Analysis of Object Detection Metrics with a Companion Open-Source Toolkit}},
  year = {2021},
  doi = {10.3390/electronics10030279},
  journal = {Electronics},
  volume = {10},
  number = {3},
  pages = {279},
}

@article{pan2025intelligent,
  author = {Pan, Yuan and Zhou, Shuang-Xi and Ding, Yang},
  title = {{Intelligent recognition of pavement crack based on RM-YOLOv5s algorithm}},
  year = {2025},
  doi = {10.1016/j.cscm.2025.e05420},
  journal = {Case Studies in Construction Materials},
  volume = {23},
  pages = {e05420},
  eid = {e05420},
}

@misc{peng2024dfine,
  author = {Peng, Yansong and Li, Hebei and Wu, Peixi and Zhang, Yueyi and Sun, Xiaoyan and Wu, Feng},
  title = {{D-FINE: Redefine Regression Task in DETRs as Fine-grained Distribution Refinement}},
  year = {2024},
  doi = {10.48550/arxiv.2410.13842},
  eprint = {2410.13842v1},
  archivePrefix = {arXiv},
  howpublished = {arXiv preprint 2410.13842v1},
  url = {https://arxiv.org/abs/2410.13842v1},
}

@misc{pottier2025modular,
  author = {Pottier, Nicolas and Lau, Meng Cheng},
  title = {{A Modular Object Detection System for Humanoid Robots Using YOLO}},
  year = {2025},
  doi = {10.48550/arxiv.2510.13625},
  eprint = {2510.13625v1},
  archivePrefix = {arXiv},
  howpublished = {arXiv preprint 2510.13625v1},
  url = {https://arxiv.org/abs/2510.13625v1},
}

@article{ramos2025comprehensive,
  author = {Ramos, Leo Thomas and Sappa, Angel D.},
  title = {{A comprehensive analysis of YOLO architectures for tomato leaf disease identification}},
  year = {2025},
  doi = {10.1038/s41598-025-11064-0},
  journal = {Scientific Reports},
  volume = {15},
  number = {1},
  eid = {26890},
  pages = {26890},
}

@misc{robinson2025rfdetr,
  author = {Robinson, Isaac and Robicheaux, Peter and Popov, Matvei and Ramanan, Deva and Peri, Neehar},
  title = {{RF-DETR: Neural Architecture Search for Real-Time Detection Transformers}},
  year = {2025},
  doi = {10.48550/arxiv.2511.09554},
  eprint = {2511.09554v2},
  archivePrefix = {arXiv},
  howpublished = {arXiv preprint 2511.09554v2},
  url = {https://arxiv.org/abs/2511.09554v2},
}

@misc{sapkota2025rfdetr,
  author = {Sapkota, Ranjan and Cheppally, Rahul Harsha and Sharda, Ajay and Karkee, Manoj},
  title = {{RF-DETR Object Detection vs YOLOv12 : A Study of Transformer-based and CNN-based Architectures for Single-Class and Multi-Class Greenfruit Detection in Complex Orchard Environments Under Label Ambiguity}},
  year = {2025},
  doi = {10.48550/arxiv.2504.13099},
  eprint = {2504.13099v1},
  archivePrefix = {arXiv},
  howpublished = {arXiv preprint 2504.13099v1},
  url = {https://arxiv.org/abs/2504.13099v1},
}

@misc{sapkota2025ultralytics,
  author = {Sapkota, Ranjan and Karkee, Manoj},
  title = {{Ultralytics YOLO Evolution: An Overview of YOLO27, YOLO26, YOLO11, YOLOv8, and YOLOv5 Object Detectors for Computer Vision and Pattern Recognition}},
  year = {2025},
  doi = {10.48550/arxiv.2510.09653},
  eprint = {2510.09653v4},
  archivePrefix = {arXiv},
  howpublished = {arXiv preprint 2510.09653v4},
  url = {https://arxiv.org/abs/2510.09653v4},
}

@article{sengun2025comparative,
  author = {Sengun, Esra and Aksoy, Samet and Sertel, Elif and Fransson, Johan E. S.},
  title = {{Comparative Analysis of YOLOv8 and YOLOv11 on Tree Detection Using UAV RGB and Laser Scanning Data}},
  year = {2025},
  doi = {10.5194/isprs-annals-x-2-w2-2025-173-2025},
  journal = {ISPRS Annals of the Photogrammetry, Remote Sensing and Spatial Information Sciences},
  volume = {X-2/W2-2025},
  pages = {173--179},
}

@article{shalash2026benchmarking,
  author = {Shalash, Omar and Khatab, Esraa and El-Agamy, Ahmed and Elmokadem, Loay and Abouelsaad, Yasmin and Zaki, Jasser and El-Sayed, Mohamed and Said, Hany},
  title = {{Benchmarking YOLOv8--YOLOv12 for Real-Time Object Detection on Single-Board Computers}},
  year = {2026},
  doi = {10.3390/make8070204},
  journal = {Machine Learning and Knowledge Extraction},
  volume = {8},
  number = {7},
  pages = {204},
}

@article{siddiqui2020drone,
  author = {Siddiqui, Zahid Ali and Park, Unsang},
  title = {{A Drone Based Transmission Line Components Inspection System with Deep Learning Technique}},
  year = {2020},
  doi = {10.3390/en13133348},
  journal = {Energies},
  volume = {13},
  number = {13},
  pages = {3348},
}

@misc{sowrirajan2020mococxr,
  author = {Sowrirajan, Hari and Yang, Jingbo and Ng, Andrew Y. and Rajpurkar, Pranav},
  title = {{MoCo-CXR: MoCo Pretraining Improves Representation and Transferability of Chest X-ray Models}},
  year = {2020},
  doi = {10.48550/arxiv.2010.05352},
  eprint = {2010.05352v3},
  archivePrefix = {arXiv},
  howpublished = {arXiv preprint 2010.05352v3},
  url = {https://arxiv.org/abs/2010.05352v3},
}

@article{sundar2023meteorological,
  author = {Sundar, Srihari and Craig, Michael T. and Payne, Ashley E. and Brayshaw, David J. and Lehner, Flavio},
  title = {{Meteorological drivers of resource adequacy failures in current and high renewable Western U.S. power systems}},
  year = {2023},
  doi = {10.1038/s41467-023-41875-6},
  journal = {Nature Communications},
  volume = {14},
  number = {1},
  eid = {6379},
  pages = {6379},
}

@article{tasyurek2024dgag,
  author = {Ta{\c{s}}y{\"{u}}rek, Murat and K{\i}l{\i}{\c{c}}kaya, Fatma Nur and {\"{O}}zt{\"{u}}rk, Celal},
  title = {{DGAG: A Novel Approach That Detects Green Areas With Geolocation From Large-Scale Satellite Imaging Instance Segmentation and GIS Application}},
  year = {2024},
  doi = {10.1111/tgis.13264},
  journal = {Transactions in GIS},
  volume = {28},
  number = {8},
  pages = {2742--2762},
}

@article{tian2024performance,
  author = {Tian, Jiya and Jin, Qiangshan and Wang, Yizong and Yang, Jie and Zhang, Shuping and Sun, Dengxun},
  title = {{Performance analysis of deep learning-based object detection algorithms on COCO benchmark: a comparative study}},
  year = {2024},
  doi = {10.1186/s44147-024-00411-z},
  journal = {Journal of Engineering and Applied Science},
  volume = {71},
  number = {1},
  eid = {76},
  pages = {76},
}

@article{tiede2021mask,
  author = {Tiede, Dirk and Schwendemann, Gina and Alobaidi, Ahmad and Wendt, Lorenz and Lang, Stefan},
  title = {{Mask R-CNN-based building extraction from VHR satellite data in operational humanitarian action: An example related to Covid-19 response in Khartoum, Sudan}},
  year = {2021},
  doi = {10.1111/tgis.12766},
  journal = {Transactions in GIS},
  volume = {25},
  number = {3},
  pages = {1213--1227},
}

@article{usery2022geoai,
  author = {Usery, E. Lynn and Arundel, Samantha T. and Shavers, Ethan and Stanislawski, Lawrence and Thiem, Philip and Varanka, Dalia},
  title = {{GeoAI in the US Geological Survey for topographic mapping}},
  year = {2022},
  doi = {10.1111/tgis.12830},
  journal = {Transactions in GIS},
  volume = {26},
  number = {1},
  pages = {25--40},
}

@article{vieiraesilva2023insplad,
  author = {Vieira e Silva, Andr{\'{e}} Luiz Buarque and de Castro Felix, Heitor and Sim{\~{o}}es, Franscisco Paulo Magalh{\~{a}}es and Teichrieb, Veronica and dos Santos, Michel and Santiago, Hemir and Sgotti, Virginia and Lott Neto, Henrique},
  title = {{InsPLAD: A Dataset and Benchmark for Power Line Asset Inspection in UAV Images}},
  year = {2023},
  doi = {10.1080/01431161.2023.2283900},
  journal = {International Journal of Remote Sensing},
  volume = {44},
  number = {23},
  pages = {7294--7320},
}

@article{vrsalovic2025system,
  author = {Vrsalovic, Ivan and Lerga, Jonatan and Ivasic-Kos, Marina},
  title = {{A System for Real-Time Detection of Abandoned Luggage}},
  year = {2025},
  doi = {10.3390/s25092872},
  journal = {Sensors},
  volume = {25},
  number = {9},
  pages = {2872},
}

@article{xu2024resilience,
  author = {Xu, Luo and Feng, Kairui and Lin, Ning and Perera, A.T.D. and Poor, H. Vincent and Xie, Le and Ji, Chuanyi and Sun, X. Andy and Guo, Qinglai and O'Malley, Mark},
  title = {{Resilience of renewable power systems under climate risks}},
  year = {2024},
  doi = {10.1038/s44287-023-00003-8},
  journal = {Nature Reviews Electrical Engineering},
  volume = {1},
  number = {1},
  pages = {53--66},
}

@article{xu2025quantifying,
  author = {Xu, Luo and Lin, Ning and Poor, H. Vincent and Xi, Dazhi and Perera, A. T. D.},
  title = {{Quantifying cascading power outages during climate extremes considering renewable energy integration}},
  year = {2025},
  doi = {10.1038/s41467-025-57565-4},
  journal = {Nature Communications},
  volume = {16},
  number = {1},
  eid = {2582},
  pages = {2582},
}

@article{yang2025yolov10msa,
  author = {Yang, Junbiao and Mat Isa, Nor Ashidi},
  title = {{YOLOv10-MsA: Attention-Augmented Real-Time Insulator Defect Detection from UAV Imagery}},
  year = {2025},
  doi = {10.28991/esj-2025-09-06-02},
  journal = {Emerging Science Journal},
  volume = {9},
  number = {6},
  pages = {2899--2914},
}

@article{yao2023site,
  author = {Yao, Yao and Feng, Chenqi and Xie, Jiteng and Yan, Xiaoqin and Guan, Qingfeng and Han, Jian and Zhang, Jiaqi and Ren, Shuliang and Liang, Yuyun and Luo, Peng},
  title = {{A site selection framework for urban power substation at micro-scale using spatial optimization strategy and geospatial big data}},
  year = {2023},
  doi = {10.1111/tgis.13093},
  journal = {Transactions in GIS},
  volume = {27},
  number = {6},
  pages = {1662--1679},
}

@article{yigit2025deep,
  author = {Yi{\u{g}}it, Abdurahman Yasin},
  title = {{Deep Learning-Based Palm Tree Detection for Urban Green Space Monitoring Using High-Resolution UAV Imagery}},
  year = {2025},
  doi = {10.1111/tgis.70171},
  journal = {Transactions in GIS},
  volume = {29},
  number = {8},
  eid = {e70171},
  pages = {e70171},
}

@misc{yu2019scale,
  author = {Yu, Xuehui and Gong, Yuqi and Jiang, Nan and Ye, Qixiang and Han, Zhenjun},
  title = {{Scale Match for Tiny Person Detection}},
  year = {2019},
  doi = {10.48550/arxiv.1912.10664},
  eprint = {1912.10664v1},
  archivePrefix = {arXiv},
  howpublished = {arXiv preprint 1912.10664v1},
  url = {https://arxiv.org/abs/1912.10664v1},
}

@article{zhang2022finet,
  author = {Zhang, Zheng-De and Zhang, Bo and Lan, Zhi-Cai and Liu, Hai-Chun and Li, Dong-Ying and Pei, Ling and Yu, Wen-Xian},
  title = {{FINet: An Insulator Dataset and Detection Benchmark Based on Synthetic Fog and Improved YOLOv5}},
  year = {2022},
  doi = {10.1109/tim.2022.3194909},
  journal = {IEEE Transactions on Instrumentation and Measurement},
  volume = {71},
  pages = {1--8},
}

@article{zhang2023condition,
  author = {Zhang, Lishuo and Ma, Zhuxing and Gu, Hao and Xin, Zizhong and Han, Pengcheng},
  title = {{Condition Monitoring and Analysis Method of Smart Substation Equipment Based on Deep Learning in Power Internet of Things}},
  year = {2023},
  doi = {10.4018/ijitsa.324519},
  journal = {International Journal of Information Technologies and Systems Approach},
  volume = {16},
  number = {3},
  pages = {1--16},
}

@article{zhang2023lightweight,
  author = {Zhang, Tong and Zhang, Yinan and Xin, Min and Liao, Jiashe and Xie, Qingfeng},
  title = {{A Light-Weight Network for Small Insulator and Defect Detection Using UAV Imaging Based on Improved YOLOv5}},
  year = {2023},
  doi = {10.3390/s23115249},
  journal = {Sensors},
  volume = {23},
  number = {11},
  pages = {5249},
}

@article{zhao2023defect,
  author = {Zhao, Zhenbing and Lv, Xuechun and Xi, Yue and Miao, Siyu},
  title = {{Defect detection method for key area guided transmission line components based on knowledge distillation}},
  year = {2023},
  doi = {10.3389/fenrg.2023.1287024},
  journal = {Frontiers in Energy Research},
  volume = {11},
  eid = {1287024},
  pages = {1287024},
}

@misc{zhao2023detrs,
  author = {Zhao, Yian and Lv, Wenyu and Xu, Shangliang and Wei, Jinman and Wang, Guanzhong and Dang, Qingqing and Liu, Yi and Chen, Jie},
  title = {{DETRs Beat YOLOs on Real-time Object Detection}},
  year = {2023},
  doi = {10.48550/arxiv.2304.08069},
  eprint = {2304.08069v3},
  archivePrefix = {arXiv},
  howpublished = {arXiv preprint 2304.08069v3},
  url = {https://arxiv.org/abs/2304.08069v3},
}

@article{zheng2021component,
  author = {Zheng, Xiangyu and Jia, Rong and {Aisikaer} and Gong, Linling and Zhang, Guangru and Dang, Jian},
  title = {{Component identification and defect detection in transmission lines based on deep learning}},
  year = {2021},
  doi = {10.3233/jifs-189353},
  journal = {Journal of Intelligent \& Fuzzy Systems},
  volume = {40},
  number = {2},
  pages = {3147--3158},
}

@misc{zhu2020deformable,
  author = {Zhu, Xizhou and Su, Weijie and Lu, Lewei and Li, Bin and Wang, Xiaogang and Dai, Jifeng},
  title = {{Deformable DETR: Deformable Transformers for End-to-End Object Detection}},
  year = {2020},
  doi = {10.48550/arxiv.2010.04159},
  eprint = {2010.04159v4},
  archivePrefix = {arXiv},
  howpublished = {arXiv preprint 2010.04159v4},
  url = {https://arxiv.org/abs/2010.04159v4},
}
\end{document}